%% file: main.tex
\documentclass[]{template}

\usepackage[utf8]{inputenc}             %
\usepackage[T1]{fontenc}                %
\usepackage{url}                        %
\usepackage{booktabs}                   %
\usepackage{multirow}
\usepackage{colortbl}
\usepackage{multicol}
\usepackage{amsfonts}                   %
\usepackage{nicefrac}                   %
\usepackage{microtype}                  %
\usepackage[dvipsnames]{xcolor}         %

\usepackage{latexsym}

\usepackage{graphicx}
\usepackage{float}
\usepackage{subcaption}
\usepackage{wrapfig}
\usepackage{lipsum}
\usepackage{adjustbox}

\usepackage{bm}

\usepackage{tabularx} 
\usepackage{threeparttable}
\usepackage{ragged2e} 
\newcolumntype{L}{>{\RaggedRight\hangafter=1\hangindent=0em}X}

\usepackage{enumitem}

\usepackage{amsmath}
\usepackage{amssymb}
\usepackage{mathtools}
\usepackage{amsthm}

\setboolean{logo}{true}    %

\usepackage[linesnumbered,ruled,vlined]{algorithm2e}

\hypersetup{
    colorlinks=true,
    linkcolor=red,
    citecolor=Cerulean,
    filecolor=magenta,      
    urlcolor=magenta,
}

\usepackage[capitalize,noabbrev]{cleveref}
\crefname{section}{§}{§§}
\Crefname{section}{§}{§§}

\usepackage{calligra}
\DeclareMathAlphabet{\mathcalligra}{T1}{calligra}{m}{n}

\usepackage{pifont}

\theoremstyle{plain}

\theoremstyle{definition}

\theoremstyle{remark}

\renewcommand{\paragraph}[1]{\vspace{1mm}\noindent\textbf{#1}}

\DeclareCaptionLabelFormat{cont}{#1~#2\alph{ContinuedFloat}}
\usepackage[most]{tcolorbox}
\tcbset{
  promptbox/.style={
    top=10pt,
    colback=lightgray!20,
    colframe=Black,
    colbacktitle=NavyBlue,
    enhanced,
    center,
    attach boxed title to top center={yshift=-0.1in,xshift=0.0in},
    boxed title style={boxrule=0pt,colframe=white,},
  }
}
\newtcolorbox{promptbox}[2][]{promptbox, title=#2,#1}
\tcbset{
  takeawaybox/.style={
    top=10pt,
    colback=lightgray!20,
    colframe=Black,
    colbacktitle=BurntOrange,
    enhanced,
    center,
    attach boxed title to top center={yshift=-0.1in,xshift=0.0in},
    boxed title style={boxrule=0pt,colframe=white,},
  }
}
\newtcolorbox{takeawaybox}[2][]{takeawaybox, title=#2,#1}
\tcbset{
  observationbox/.style={
    top=10pt,
    colback=lightgray!20,
    colframe=Black,
    colbacktitle=YellowGreen,
    enhanced,
    center,
    attach boxed title to top center={yshift=-0.1in,xshift=0.0in},
    boxed title style={boxrule=0pt,colframe=white,},
  }
}
\newtcolorbox{observationbox}[2][]{observationbox, title=#2,#1}

\usepackage{xspace}

\newcommand\blfootnote[1]{%
  \begingroup
  \renewcommand\thefootnote{}\footnote{#1}%
  \addtocounter{footnote}{-1}%
  \endgroup
}

\usepackage{CJK}

\title{Intern-S2-Preview: Scientific Agentic Foundation Model}

\author[]{Intern-S2-Preview Team, Shanghai AI Laboratory}

\input{sections/0.abstract}

\begin{document}

\blfootnote{$*$ Model is available at \url{https://huggingface.co/internlm/Intern-S2-Preview}}

\maketitle

\input{sections/1.introduction}
\input{sections/2.arch}

\input{sections/3.pretrain}
\input{sections/4.posttrain}
\input{sections/5.evaluation}
\input{sections/6.conclusion}

\clearpage
\bibliographystyle{plain}
\bibliography{refs}

\clearpage

\end{document}

%% file: sections/0.abstract.tex
\begin{abstract}
Scientific discovery increasingly requires AI systems that can reason over scientific evidence of heterogeneous modalities, interact with scientific tools and environments, and sustain progress across long task horizons. We present Intern-S2-Preview, a series of scientific agentic foundation models designed to support multimodal scientific understanding, reasoning, generation, and long-horizon tasks. The training pipeline begins with scientific multimodal pre-training over rendered scientific documents, interleaved image-text data, and diverse scientific corpora. Starting from the pretrained checkpoint, we apply a unified post-training pipeline consisting of supervised fine-tuning, scalable multi-task reinforcement learning (RL), black- and white-box agentic RL, and on-policy distillation. This pipeline is supported by practical techniques that improve rollout and training stability and efficiency, including partial rollout with off-policy correction, adaptive length regularization, online speculative decoding, robust multi-task optimization, and trace-aware experience assembly for agentic tasks.
At the architecture level, Intern-S2-Preview-397B extends time series modelling from efficient long-sequence understanding to numerical forecasting, while Memory Decoder is studied as a separate memory-augmented path for rapid scientific specialization without modifying the frozen 397B backbone.
Evaluations across scientific, multimodal, agentic, and general-purpose benchmarks show that Intern-S2-Preview-397B achieves competitive or leading results in multiple settings. The time series modules improve scientific signal understanding and forecasting on SciTS, while the separate Intern-MemDec-4B extension improves the Biology-Instructions average score from 56.92 to 60.32 without modifying the frozen 397B backbone.
\end{abstract}

%% file: sections/1.introduction.tex
\section{Introduction}
Recent advances in large language models and multimodal foundation models are reshaping the development of AI for Science, enabling models to reason over diverse scientific knowledge, observations, and computational tools~\cite{taylor2022galactica,hu2025survey,bai2025intern,zou2026intern}. Scientific multimodal models and benchmarks further extend this direction beyond text by evaluating perception, understanding, and reasoning over scientific figures, microscopy images, remote-sensing observations, earth-science phenomena, and numerical time series~\cite{zhou2025scientistsexamprobingcognitive,burgess2025microvqamultimodalreasoningbenchmark,Wang_2025_CVPR,zhao2025msearthmultimodalscientificdataset,wu2025scits}. However, meaningful scientific discovery involves more than producing a correct response to an isolated question. It requires sustained reasoning and adaptive planning based on heterogeneous evidence, and repeated interaction with tools and external environments over long task horizons~\cite{tang2026sciexplore,wang2025scireasonerlayingscientificreasoning}.

Existing model families remain incomplete for such scientific workflows. General-purpose LLMs~\cite{achiam2023gpt,openai2026gpt5systemcard} provide broad instruction following and reasoning abilities, but are not specialized for heterogeneous scientific modalities, domain protocols, or verifiable tool interaction. Scientific multimodal models~\cite{bai2025intern,zou2026intern} improve perception and reasoning over specialized inputs, but are still often evaluated as static question-answering systems rather than long-horizon agents. These limitations motivate Intern-S2-Preview, a series of scientific agentic foundation models designed to move beyond scientific question answering toward iterative, tool-grounded problem solving, with Intern-S2-Preview-397B as the main model evaluated in this report.

At the architecture level, we focus on two complementary requirements for scientific agentic foundation models. First, scientific workflows often require models to both understand long numerical signals and forecast future system states. Intern-S2-Preview-397B therefore extends time series modelling from efficient long-sequence understanding to numerical forecasting by adding a dedicated forecasting branch. Second, fast adaptation to new scientific domains is often required, where the model should be specialized to a new domain without losing its general-purpose capabilities. To address this,
we explore a strategy for efficient model specialization without rewriting the model parameters, where independently trained parametric memories~\cite{wang2026memsft,wei2026memory} are attached to the frozen 397B backbone to introduce additional domain knowledge and specialized capabilities.

Intern-S2-Preview is then trained through a staged pipeline. During continual pre-training, we focus on scientific documents and multimodal corpora whose information is distributed across text, figures, tables, equations, and page layout. Visual Pre-training~\cite{Zhao2026MAPLE,zhang2026VP} learns from rendered scientific pages by predicting visual latents, allowing the model to absorb document structure that is often weakened by text extraction. In parallel, we construct interleaved PDF data by parsing pages, cropping visually informative units, and restoring text and visual elements into layout-aware sequences, with visual-gain filtering used to retain pages whose visual content contributes to language modelling. We further build a large-scale image retrieval pipeline to recall and rerank high-quality scientific images for multimodal training. Together, these stages provide the pretrained model with scientific text, document-level visual context, and cross-modal image evidence.

Starting from the pretrained checkpoint, post-training converts these pretrained capabilities into controllable reasoning, generation, and agentic behavior. Supervised fine-tuning provides the instruction-following and tool-use initialization for subsequent reinforcement learning. We then apply scalable multi-task reinforcement learning under verifiable objectives to improve reasoning depth, correctness, scientific generation, and response efficiency across heterogeneous scientific and general-purpose tasks. This stage is supported by systems and optimization techniques designed for long rollouts and heterogeneous task mixtures, including partial rollout with off-policy correction, adaptive length regularization, online speculative decoding, and Group-level Entropy-Controlled Policy Optimization (GEPO)~\cite{cheng2026groupentropycontrolledpolicyoptimization} for balancing exploration and update strength across task groups with different entropy regimes.

For long-horizon agentic tasks, we introduce a black- and white-box agentic RL framework\footnote{\url{https://github.com/InternLM/xtuner}} based on a \emph{harness $\times$ task} abstraction. The framework decouples agent runtimes from executable task distributions and aligns semantic action--observation trajectories with token-level rollout traces, so that different tool-using agents and executable tasks can share a common rollout, verification, and training protocol. We construct tasks from coding and terminal benchmarks as well as a self-evolving generalized task-synthesis system~\cite{tang2026skill2task} based on diverse community skills. Finally, on-policy distillation consolidates the separately optimized reasoning and agentic expert policies into the unified Intern-S2-Preview model.

We evaluate Intern-S2-Preview-397B across scientific, multimodal, agentic, general-purpose, and time-series benchmarks. These evaluations cover both static scientific problem solving and workflow-oriented settings that require planning, tool use, and iterative execution. The results indicate that Intern-S2-Preview-397B combines general understanding, domain-specific scientific reasoning, scientific generation, and agentic interaction within a single foundation model. It obtains competitive or leading scores on multiple scientific benchmarks, competitive open-model results on general and multimodal tasks, measurable gains on time-series understanding and forecasting, and competitive performance on agentic coding, terminal, and research-oriented tasks. We also evaluate the separate Memory Decoder variant in biology to examine modular specialization without modifying the 397B backbone.

%% file: sections/2.arch.tex
\section{Architecture}

Intern-S2-Preview-397B extends time series modelling from scientific signal understanding to numerical forecasting through upgraded time series modules. Separately, Memory Decoder provides a memory-augmented specialization path in which external parametric memories can be attached to the frozen 397B backbone without modifying the model's core parameters.

\subsection{Memory Decoder}
\label{sec:memory-decoder}

\begin{figure}[t]
\centering
\includegraphics[width=0.98\linewidth]{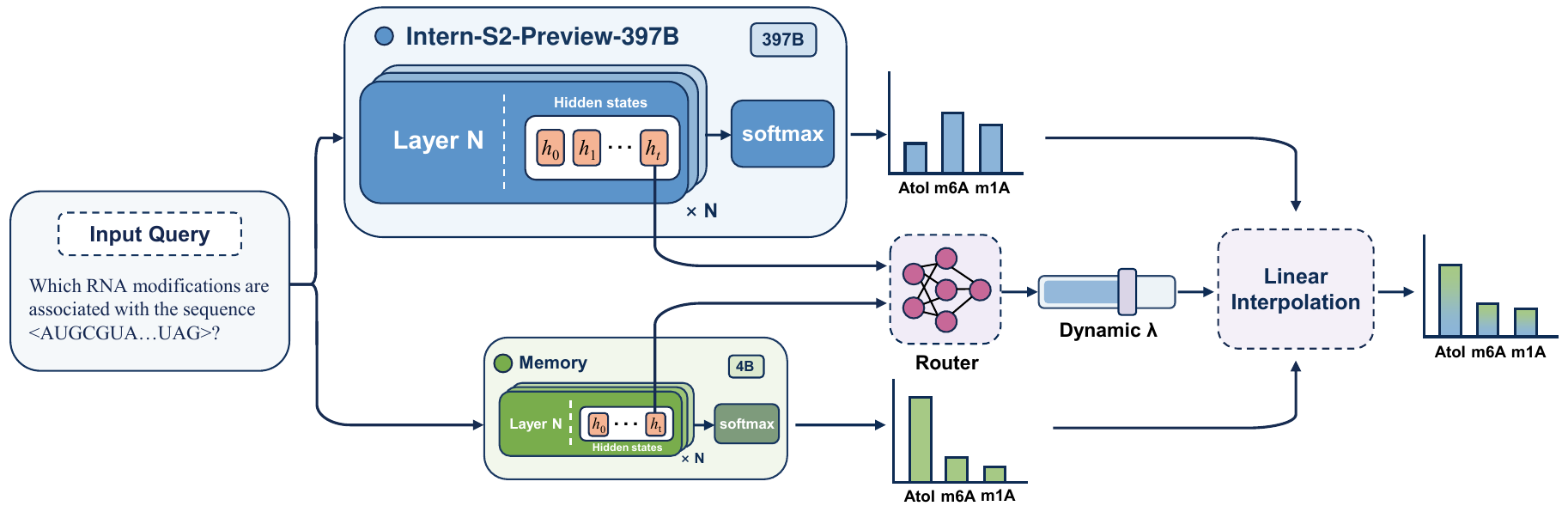}
\caption{Architecture of the separate Memory Decoder extension for Intern-S2-Preview-397B. The frozen Intern-S2-Preview-397B backbone and a domain memory process the same input in parallel and produce separate next-token distributions. A lightweight token-level router uses their hidden states and output-distribution uncertainty features to predict a dynamic fusion weight $\lambda$, which controls the contribution of the two distributions to the final prediction.}
\label{fig:intern-s2-memory-decoder-architecture}
\end{figure}

Memory Decoder~\cite{wang2026memsft,cao2026memory,wei2026memory,wei2026mlpmemory} is a separate extension model for continual domain specialization, rather than a component of the base Intern-S2-Preview-397B model. As illustrated in Figure~\ref{fig:intern-s2-memory-decoder-architecture}, it attaches new knowledge and specialized capabilities through external parametric memories while keeping the Intern-S2-Preview-397B backbone frozen. In this design, a separately trained memory decoder complements the backbone with domain-specific knowledge and capabilities through dynamic fusion of their next-token distributions. At each decoding step, a lightweight token-level router determines how much the memory decoder should contribute~\cite{wang2026memsft}. New scientific capabilities can be introduced by attaching independently trained memories without modifying the Intern-S2-Preview-397B backbone.

This design is motivated by the long-tailed and continuously evolving nature of scientific expertise. Although Intern-S2-Preview-397B provides a strong general foundation for scientific reasoning, instruction following, multimodal understanding, and tool-augmented problem solving, no fixed post-trained checkpoint can fully cover every specialized subfield, task protocol, or newly emerging domain. Directly fine-tuning the backbone for each new domain is undesirable, because the same parameter updates that improve domain performance may perturb the model's general reasoning, agentic behavior, and multimodal capabilities. Memory Decoder avoids this trade-off by turning domain extension from backbone rewriting into modular memory attachment. Intern-S2-Preview-397B continues to serve as the general-purpose backbone, while domain knowledge is supplied through plug-and-play memories without compromising general capabilities.

\paragraph{Training.}
Memory Decoder is trained by compressing retrieval-based domain evidence into a reusable parametric module. Given a domain SFT corpus $\mathcal{D}_{\mathrm{sft}}=\{(q^{(i)},a^{(i)})\}_{i=1}^{N}$, we build a token-level datastore over answer side positions. For each target token, the prefix is $c_t^{(i)}=[q^{(i)};y_{<t}^{(i)}]$, the key is $k_t^{(i)}=\phi(c_t^{(i)})$, and the value is $y_t^{(i)}$, where $\phi(\cdot)$ is frozen. Nearest neighbor retrieval over this datastore provides a soft next-token teacher distribution~\cite{khandelwal2019generalization}:
\begin{equation}
\label{eq:memory-retrieval-teacher}
p_{\mathrm{ret}}(y\mid c_t) \propto
\sum_{(k_j,v_j)\in\mathcal{N}(k_t)}
\mathbb{I}_{y=v_j}\exp(-d(k_t,k_j)/\tau),
\end{equation}
where $\mathcal{N}(k_t)$ is the retrieved neighbor set, $d(\cdot,\cdot)$ is the retrieval distance, and $\tau$ is a temperature parameter. Memory training combines retrieval distillation with supervision from the gold answer token:
\begin{equation}
\label{eq:memory-decoder-loss}
\begin{gathered}
\mathcal{L}_{\mathrm{mem}}(c_t)
=
\beta\,\mathcal{L}_{\mathrm{KL}}(c_t)
+
(1-\beta)\mathcal{L}_{\mathrm{CE}}(c_t),\\[-1pt]
\mathcal{L}_{\mathrm{KL}}(c_t)
=
\mathrm{KL}(p_{\mathrm{ret}}(\cdot\mid c_t)\|p_{\mathrm{mem}}(\cdot\mid c_t)),
\quad
\mathcal{L}_{\mathrm{CE}}(c_t)
=
-\log p_{\mathrm{mem}}(y_t\mid c_t).
\end{gathered}
\end{equation}
Here $\beta\in[0,1]$ balances the retrieval teacher and the gold SFT answer. Through this objective, the Memory Decoder learns to capture domain knowledge and recurring task patterns as a plug-and-play parametric memory.

\paragraph{Inference.}
At inference time for a memory-augmented variant, Intern-S2-Preview-397B and Memory Decoder process the same decoding context in parallel. For a prefix $c_t=[x;y_{<t}]$, the frozen Intern-S2-Preview-397B backbone produces $p_{\mathrm{S2}}(\cdot\mid c_t)$, while the memory decoder produces $p_{\mathrm{mem}}(\cdot\mid c_t)$. A lightweight token-level router takes the hidden representations from both models together with confidence and entropy features, and predicts a fusion coefficient $\lambda_t\in[0,1]$. The final next-token distribution is
\begin{equation}
\label{eq:memory-decoder-fusion-detailed}
p_{\mathrm{final}}(\cdot\mid c_t)
=
(1-\lambda_t)p_{\mathrm{S2}}(\cdot\mid c_t)
+
\lambda_t p_{\mathrm{mem}}(\cdot\mid c_t).
\end{equation}
During router training, Intern-S2-Preview-397B and Memory Decoder remain frozen, and only the router is optimized on a mixture of domain and general instruction data. In addition to cross-entropy on the fused distribution, we apply a signed linear regularizer to the memory weight:
\begin{equation}
\label{eq:memory-router-loss}
\begin{gathered}
\mathcal{L}_{\mathrm{CE}}(c_t)
=
-\log p_{\mathrm{final}}(y_t\mid c_t),
\qquad
\mathcal{R}(c_t)
=
s_t\lambda_t,\\
\mathcal{L}_{\mathrm{router}}(c_t)
=
\mathcal{L}_{\mathrm{CE}}(c_t)
+
\alpha_s\mathcal{R}(c_t).
\end{gathered}
\end{equation}
where $s_t<0$ for domain examples, $s_t>0$ for general examples, and $\alpha_s>0$ controls the regularization strength.

\begin{figure}[t]
    \centering
    \begin{subfigure}[t]{0.45\linewidth}
        \centering
        \includegraphics[width=\linewidth]{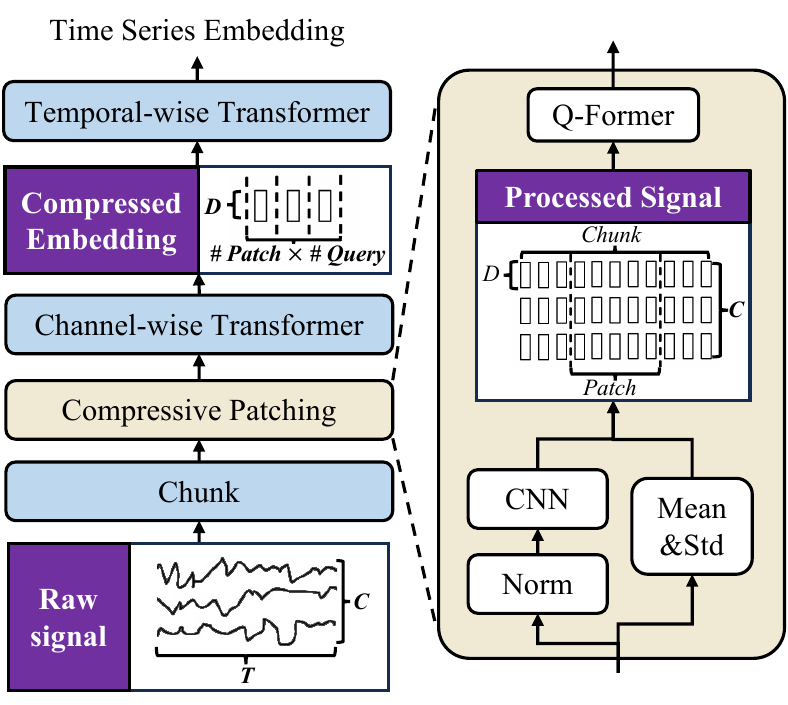}
        \caption{Structure of the time series encoder.}
        \label{fig:overall}
    \end{subfigure}
    \begin{subfigure}[t]{0.35
\linewidth}
        \centering
        \includegraphics[width=\linewidth]{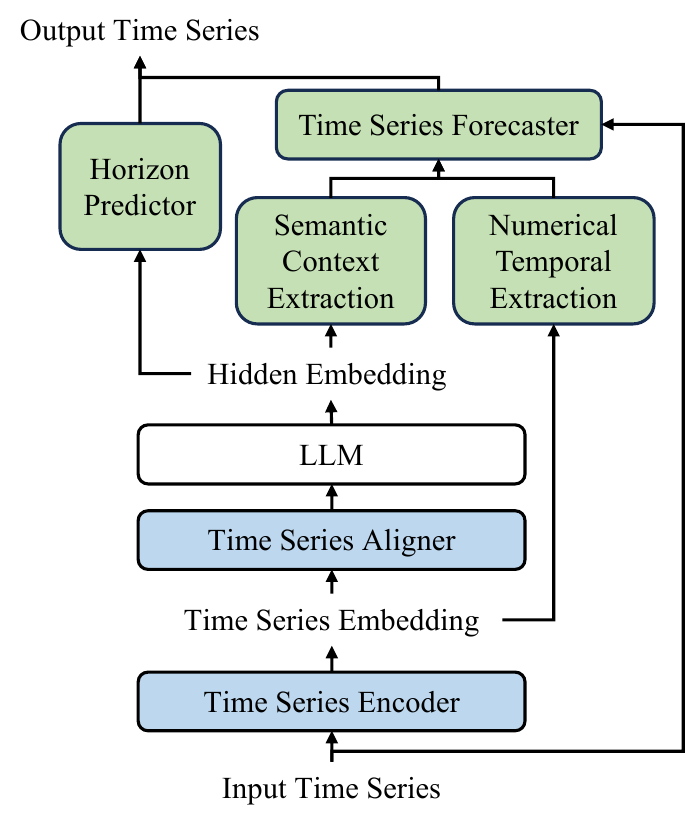}
        \caption{Structure of the time series forecaster.}
        \label{fig:forecaster}
    \end{subfigure}
    \caption{Architecture of the time series modules for long-sequence understanding and numerical forecasting.}
    \label{fig:ts_module}
\end{figure}

\subsection{Time Series Modules}
\subsubsection{Upgraded Time Series Encoder for Efficient Long-Sequence Modelling}

Scientific time series often exhibit substantial variations in sequence length, sampling frequency, and channel dependency, making efficient and expressive modelling challenging. Intern-S2-Preview-397B upgrades the time series encoder over Intern-S1-Pro with improved long-sequence processing efficiency and enhanced multi-channel representation learning.

As illustrated in Figure~\ref{fig:overall}, the input time series is first partitioned into temporal chunks, enabling localized processing of long sequences. Each chunk is processed by a compressive patching module, which consists of normalization, CNN-based local feature extraction, and Q-Former based temporal compression. During normalization, channel-wise mean and standard deviation are retained as auxiliary statistics. The extracted local representations are then divided into temporal patches, where each patch is compressed by a Q-Former with learnable queries into a fixed number of tokens. By dynamically adjusting the temporal patching process according to input length, the encoder maintains a controllable output sequence length for heterogeneous long time series. Compared with the time series encoder in Intern-S1-Pro, which directly aggregated multi-channel representations through mean pooling, Intern-S2-Preview-397B introduces a channel-wise Transformer encoder to model inter-channel dependencies before being fed into the Transformer encoder body for global temporal context modelling. The connection between the time series encoder and the LLM remains unchanged.

Compared with the previous version used in Intern-S1-Pro, the upgraded encoder increases the maximum supported input length from approximately 240,000 to 300,000 time steps. At the maximum sequence length, it achieves approximately $5\sim 6\times$ faster inference while reducing GPU memory consumption to around 20\% of the previous version. Beyond improving the processing of long sequences, the new architecture also enables effective modelling of signals with high-frequency but short sequence lengths, a setting not supported by Intern-S1-Pro. This capability further expands the disciplinary coverage of the time series module, extending its existing support for astronomy, geoscience, neuroscience, physiological signal analysis, and bioacoustics to include radar signal analysis ($\sim$MHz).

\subsubsection{Time Series Generation Module}
Beyond time series understanding, time series forecasting is essential for scientific applications as it enables models to predict future system states and support a broader range of scientific tasks. Intern-S2-Preview-397B integrates a time series forecasting module to enable unified time series understanding and generation within the multimodal LLM framework. By introducing a dedicated numerical forecasting branch rather than generating values as discrete text tokens, the model preserves numerical fidelity while maintaining computational efficiency.

As illustrated in Figure~\ref{fig:forecaster}, the forecasting module introduces a forecasting branch conditioned on multimodal representations from the LLM and the time series encoder. Semantic context from the LLM and numerical temporal representations from the time series encoder are selectively extracted by Q-Former and integrated to condition a causal Transformer forecaster via cross-attention for future sequence generation. A horizon predictor further interprets the forecasting instruction and determines the required prediction length, enabling flexible forecasting across different horizons.

%% file: sections/3.pretrain.tex
\section{Pre-training}

Scientific corpora contain knowledge in both textual and visual forms. Beyond scaling text tokens, Intern-S2-Preview strengthens its scientific data foundation by preserving document layout, linking visual units with surrounding scientific context, and retrieving high-quality visual samples for multimodal training.

\subsection{Visual Pre-training}

\begin{figure}[t]
    \centering
    \includegraphics[width=0.9\linewidth]{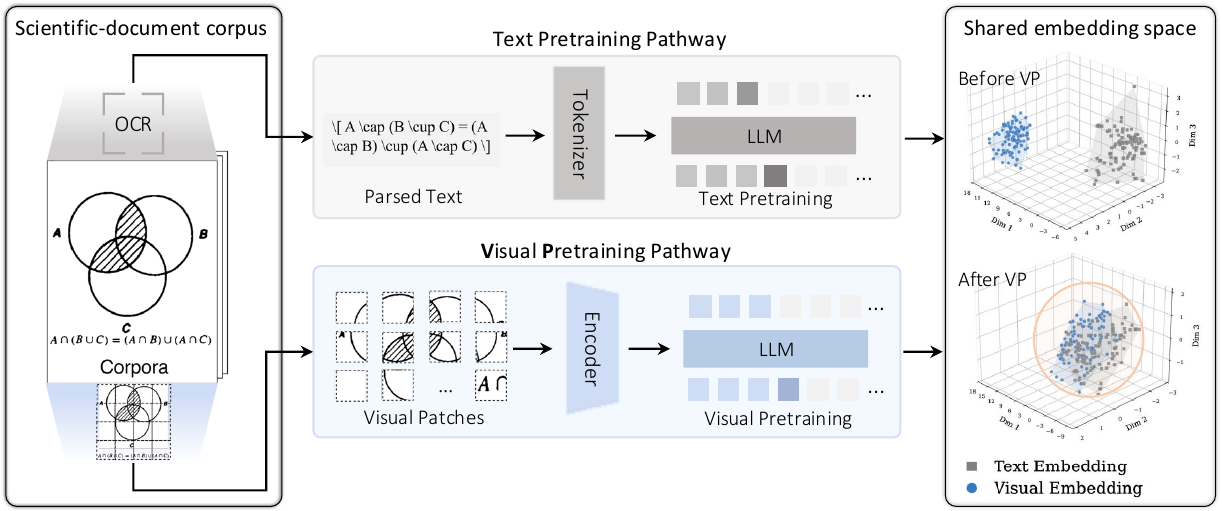}
    \caption{Overview of matched text and visual pre-training. The text pathway predicts tokens from parsed PDF content, whereas the visual pathway predicts foreground visual latents from rendered pages, improving alignment between textual and visual document representations.}
    \label{fig:vp_pipeline}
\end{figure}

Beyond text-centric scientific training, Intern-S2-Preview introduces Visual Pre-training (VP) as a lightweight stage for modality expansion. Following \cite{Zhao2026MAPLE,zhang2026VP}, VP learns from large-scale unlabeled scientific documents rendered as page images, preserving figures, tables, equations, and layout information that may be lost during text extraction. As illustrated in Figure~\ref{fig:vp_pipeline}, VP complements conventional text pre-training by learning directly from the visual representation of the same document corpus.

Given a page image $\mathcal{I}$, a frozen visual encoder extracts a sequence of visual features
$\mathcal{Z}=E_{\mathrm{v}}(\mathcal{I})=(z_1,\ldots,z_N)$. A foreground mask $m_i$ removes blank regions, after which the retained features are arranged in raster-scan order. The resulting sequence is projected into the LLM hidden space and modeled autoregressively:

\begin{equation}
\mathcal{U}
=
\operatorname{RasterScan}\{z_i \mid m_i=1\}
=
(u_1,\ldots,u_L),
\qquad L \leq N,
\end{equation}

\begin{equation}
\hat{u}_{t+1}
=
\psi\!\left(
\left[
\Phi_{\theta}
\left(
W_{\mathrm{in}}u_{\leq t}
\right)
\right]_t
\right).
\label{eq:intern_s2_vp_forward}
\end{equation}

Here, $W_{\mathrm{in}}$ denotes the visual input projection, $\Phi_{\theta}$ is the autoregressive LLM backbone, and $\psi$ is a lightweight visual prediction head. VP is trained with a contrastive next-latent prediction objective. For prediction and target indices $t,j\in\mathcal{B}$, their temperature-scaled cosine similarity and matching probability are

\begin{equation}
s_{tj}
=
\frac{
\hat{u}_{t+1}^{\top}u_{j+1}
}{
\tau
\lVert \hat{u}_{t+1}\rVert_2
\lVert u_{j+1}\rVert_2
},
\qquad
p_{tj}
=
\frac{\exp(s_{tj})}
{\sum_{k\in\mathcal{B}}\exp(s_{tk})}.
\label{eq:intern_s2_vp_probability}
\end{equation}

The VP loss is defined as

\begin{equation}
\mathcal{L}_{\mathrm{VP}}
=
-\frac{1}{|\mathcal{B}|}
\sum_{t\in\mathcal{B}}
\log p_{tt},
\label{eq:intern_s2_vp_loss}
\end{equation}

where $u_{t+1}$ is the positive target of $\hat{u}_{t+1}$, and the remaining targets in $\mathcal{B}$ serve as in-batch negatives. During continued pre-training, text and visual samples are interleaved under the joint objective

\begin{equation}
\mathcal{L}
=
\lambda_{\mathrm{text}}\mathcal{L}_{\mathrm{CE}}
+
\lambda_{\mathrm{vis}}\mathcal{L}_{\mathrm{VP}}.
\label{eq:intern_s2_vp_objective}
\end{equation}

The visual encoder remains frozen, while the LLM backbone, visual projection, and prediction head are optimized. Since supervision is obtained directly from visual features, VP requires neither OCR and layout parsing nor paired data and manual annotations. It therefore provides a scalable complement to text pre-training, retaining document structures and visual patterns that support both language and multimodal scientific capabilities.

\begin{figure}[t]
    \centering
    \includegraphics[width=1.0\linewidth]{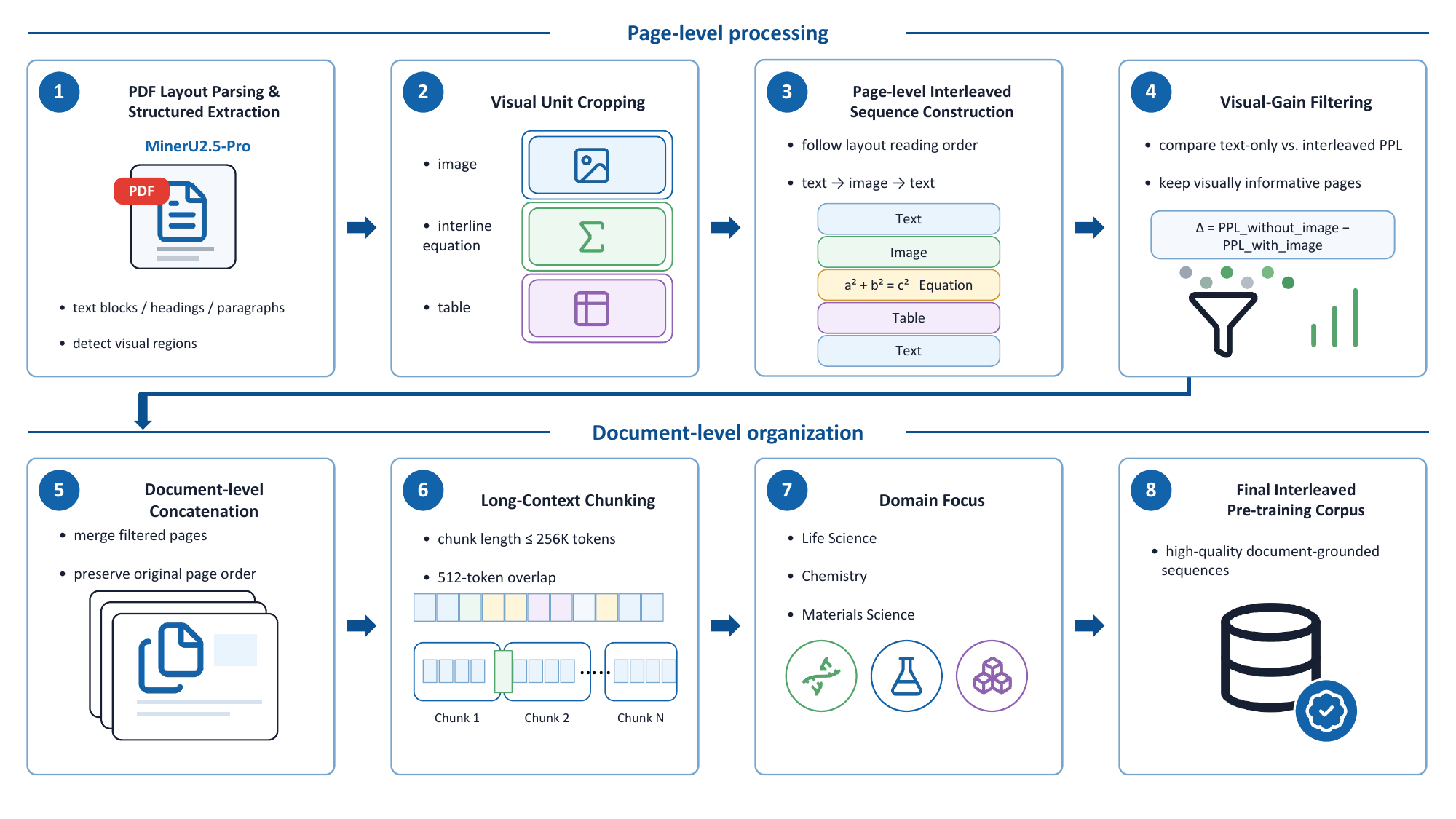}
    \caption{Pipeline for producing the interleaved image-text pair data from PDF documents, including OCR and layout parsing, visual-unit cropping, visual-gain filtering, and document-level sequence assembly.}
    \label{fig:pretrain_interleaved_data}
\end{figure}

\subsection{Interleaved Text-Image Data}

In pre-training of a multi-modal large model, constructing image-text pairs alone is insufficient to fully cover the multi-modal understanding demands of real-world PDF documents. Existing multimodal pre-training strategies heavily rely on image caption data, which captures semantic alignment between an image and a local text span, making them suitable for object recognition, image description, and local visual-semantic modeling. However, the more critical information in PDFs often lies in the contextual relationships between images, equations, tables, and surrounding text, including layout position, explanatory paragraphs before and after visual elements, textual references, cross-page reasoning chains, and the progressive organization of knowledge in long documents. Therefore, we further construct interleaved image-text data from PDFs, as shown in Figure~\ref{fig:pretrain_interleaved_data}, enabling the model to learn not only what an image depicts, but also how it is embedded in document narratives and participates in knowledge expression and reasoning.

Specifically, we first apply MinerU2.5-Pro \cite{wang2026mineru25propushinglimitsdatacentric} to perform OCR and layout-aware structural parsing on PDF documents. The system identifies text blocks, headings, paragraphs, and visually informative regions on each page. For visual content, we focus on three types of units: regular images, interline equations, and tables. Each visual unit is cropped from the original PDF page according to its bounding box and saved as a standardized sub-image. We then construct page-level interleaved sequences. For each page, text blocks and visual units are reorganized according to the layout reading order and bounding-box order, forming page-level sequences. To further select high-value pages with genuine visual dependency, we introduce a visual-gain-based quality filtering mechanism. Inspired by Toolformer \cite{schick2023toolformerlanguagemodelsteach}, we compute the language model perplexity of the page text under two conditions: a text-only condition without visual inputs, and an interleaved condition with images, tables, or equations included. Visual gain is defined as the difference between the two perplexities. A significant decrease in PPL after adding visual information indicates that the visual content provides meaningful support for understanding the page. Decorative images, advertisements, or weakly related visual elements typically yield low visual gain, whereas scientific pages containing experimental figures, mechanism diagrams, structural illustrations, tables, or equations often lead to a notable PPL reduction. Therefore, we combine human review with domain-specific thresholds and retain only pages whose visual gain exceeds the corresponding threshold.

Finally, we concatenate the filtered page-level interleaved sequences in the original PDF page order to form document-level sequences, which are further split into chunks suitable for long-context VLM pre-training, with each chunk capped at 256k tokens and sharing a 512-token overlap. This pipeline focuses on life sciences, chemistry, and materials science, yielding high-quality interleaved image-text data for VLM pre-training.

\begin{figure}[t]
    \centering
    \includegraphics[width=1.0\linewidth]{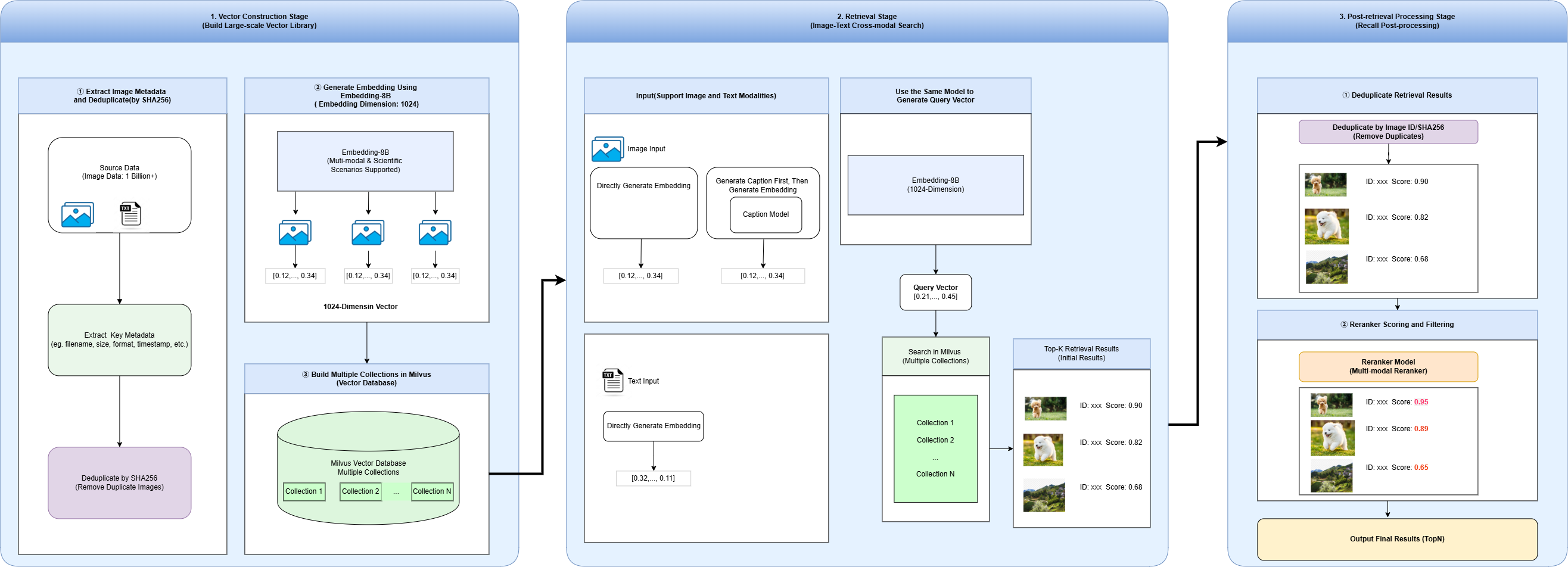}
    \caption{The pipeline of the image retrieval process, including image encoding, vector database construction, and online text-to-image and image-to-image retrieval with post-processing.}
    \label{fig:pretrain_image_retrieval}
\end{figure}

\subsection{Image Retrieval Enhancement}

Retrieval of high-quality data is a common practice in preparing textual pre-training corpus. However, the pipeline of retrieving high-quality image data is underexplored. Thus, as shown in Figure~\ref{fig:pretrain_image_retrieval}, we introduce a large-scale image retrieval pipeline to recall high-quality data and raise their sample ratio during the training to enhance the model's multimodal ability. 

The pipeline relies on building a large-scale image vector database. The main process includes: 1) extracting images and their key metadata from delivered data sources, and deduplicating them according to the SHA256 values of images to ensure the uniqueness of images in the vector database; 2) using an 8B embedding model to encode images and generate 1024-dimensional embedding representations; 3) in order to balance storage and retrieval performance for data at the scale of hundreds of millions, constructing multiple collections based on the Milvus vector database and storing image vectors in shards to support subsequent high-performance retrieval.

\paragraph{Online retrieval stage.} The system supports two retrieval modes: text-to-image retrieval and image-to-image retrieval. For both modes, the same embedding model as used in the vector construction stage is adopted for vector encoding, so as to maintain consistency in the vector space.

For different types of input, the processing flow is as follows. 
\begin{itemize}
    \item \textbf{Image input:} the input image is directly encoded to obtain its image embedding; meanwhile, a caption model \cite{xing2026caprl} is used to generate a textual description of the image, and the caption text is then encoded into a vector. In this way, joint retrieval is performed from both visual and semantic perspectives, which improves recall and semantic matching ability.
    \item \textbf{Text input:} the embedding model is directly used to generate the text vector, and cross-modal similarity retrieval is performed in the image vector database.
\end{itemize}

\paragraph{Post-processing stage after recall.} To further improve the quality of retrieved results, the system applies post-processing to recalled results, including: filtering duplicate samples, using a reranker model to rerank candidate results and assign quality scores, and utilizing the scores to finally filter retrieval results.

%% file: sections/4.posttrain.tex
\section{Post-Training}

\subsection{Post-Training Framework}

Starting from the pretrained checkpoint, we develop a unified post-training pipeline for Intern-S2-Preview. The pipeline strengthens general reasoning, instruction following, tool use, and long-horizon agentic behavior, while further improving three core capabilities for scientific intelligence: scientific reasoning, generation across scientific modalities, and scientific agentic problem solving.

\begin{figure}[t]
    \centering
    \includegraphics[width=\linewidth]{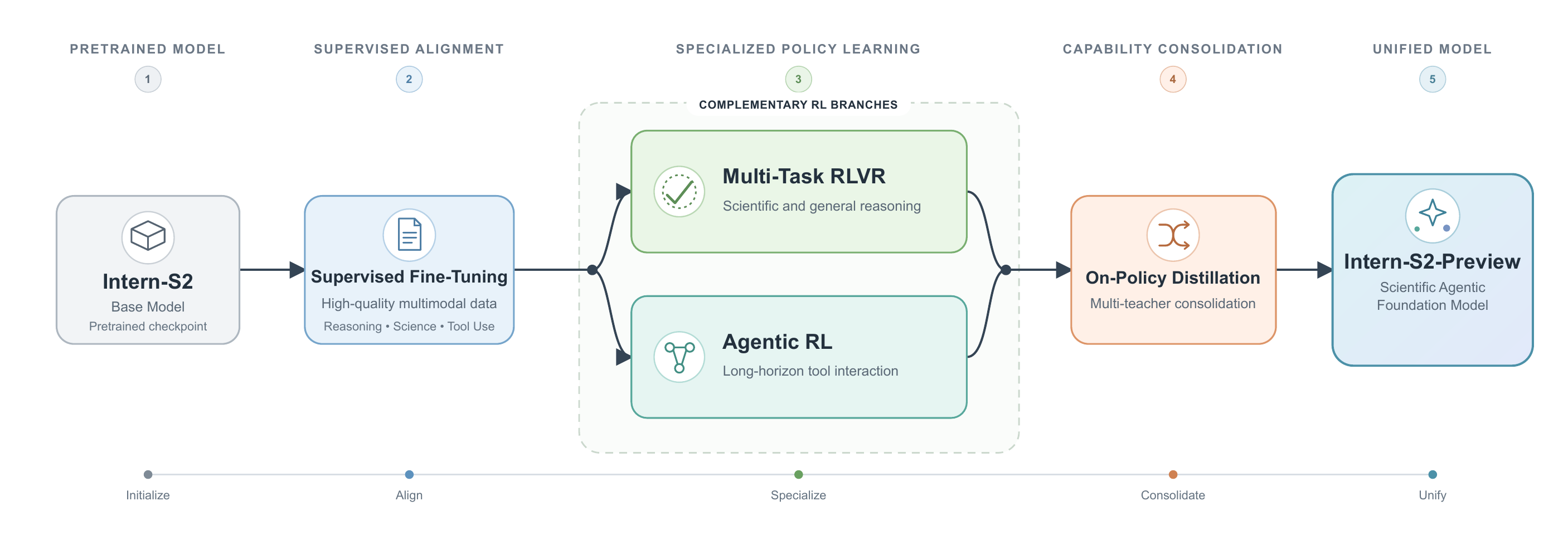}
    \caption{
        Overview of the post-training pipeline for Intern-S2-Preview.
        The pretrained base model is first enhanced through supervised fine-tuning, followed by multi-task RLVR and black-box agentic RL for general and specialized capability improvement. 
        On-policy distillation then consolidates the resulting scientific reasoning and agentic capabilities into a single unified model.
    }
    \label{fig:post_training_pipeline}
\end{figure}

As illustrated in Figure~\ref{fig:post_training_pipeline}, the pipeline consists of three major stages. We first perform supervised fine-tuning on a broad mixture of high-quality demonstrations to establish fundamental reasoning behaviors, response formats, scientific generation capabilities, and tool-use patterns.
We then conduct scalable multi-task reinforcement learning over diverse scientific and general-purpose tasks to further improve reasoning, generation, and scientific capabilities. 
In parallel, we apply black-box agentic reinforcement learning to cultivate specialized policies in interactive environments, where the model solves complex tasks through external tools and environmental feedback. 
Finally, we employ on-policy distillation to consolidate the capabilities acquired by the general and specialized policies into a single unified model.

\subsection{Supervised Fine-Tuning}

The first post-training stage converts the pretrained model into a controllable assistant before applying reinforcement learning.
We perform supervised fine-tuning on a large-scale, high-quality multimodal dataset covering a broad range of domains and interaction settings. 
The data mixture includes general conversation, instruction following, safety alignment, code generation and reasoning, image--text understanding, visual perception and spatial grounding, tool use, specialized scientific tasks, and long-horizon agentic trajectories. 
This diverse supervision equips the model with strong foundational capabilities across both general-purpose and scientific scenarios.

To ensure data quality, we apply extensive filtering, cleaning, and deduplication procedures. 
For tasks requiring explicit reasoning, we construct high-quality chain-of-thought demonstrations through rejection sampling using our previous-generation model, Intern-S1-Pro, together with other leading open-source models. 
The resulting samples are further validated by language models and human domain experts to improve factual correctness, reasoning quality, and format consistency. 
This carefully curated SFT stage provides a strong and stable initialization for the subsequent reinforcement learning stages.

\subsection{Scalable and Stable Reinforcement Learning}

After SFT, reinforcement learning is used to improve correctness, reasoning depth, scientific generation, and response efficiency under verifiable objectives.
Scaling the reinforcement learning pipeline introduces several practical challenges. Rollout generation is the primary computational bottleneck in reinforcement learning.
Partial and asynchronous rollouts also introduce off-policy effects that require careful control. 
In addition, balancing reasoning efficiency with model performance remains difficult, while optimization conflicts in multi-task training can hinder stable convergence across domains. To address these challenges, we introduce partial rollout with off-policy correction, adaptive length regularization, speculative decoding for faster RL rollouts, and robust multi-task optimization.
The following sections describe the individual components of our post-training framework in detail.

\subsubsection{Efficient Partial Rollout with Off-Policy Correction}

Long-reasoning reinforcement learning is particularly susceptible to the long-tail distribution of response lengths. In a synchronous rollout pipeline, a small number of exceptionally long generations may delay the completion of an entire batch, leaving most GPUs idle while waiting for stragglers~\cite{zhou2025april,qu2025copris,fu2025areal}. Existing systems commonly address this problem either by co-locating training and inference with partial rollouts~\cite{sheng2025hybridflow,zhou2025april,qu2025copris}, or by fully disaggregating rollout generation and policy optimization onto separate GPU pools~\cite{fu2025areal}.

\begin{figure}[t]
    \centering
    \includegraphics[width=\linewidth]{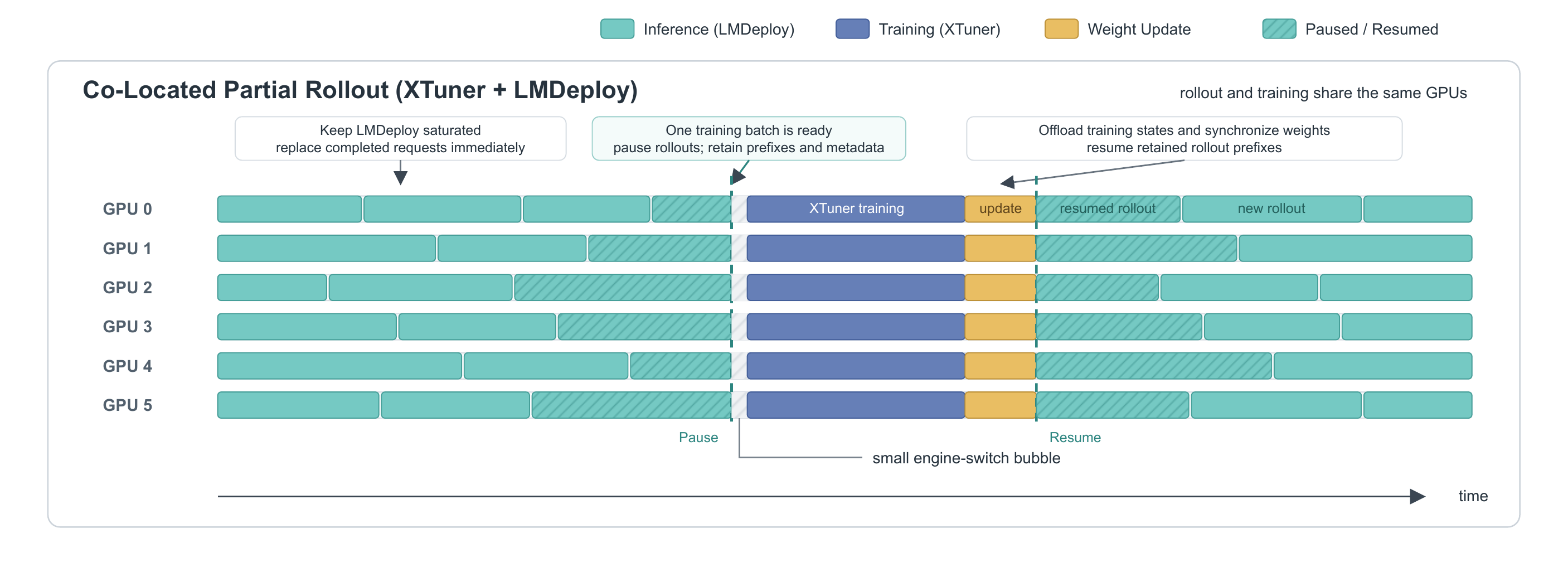}
    \caption{Overview of our co-located partial-rollout system based on XTuner and LMDeploy. During rollout, completed requests are continuously replaced to keep the inference engine fully utilized. Once sufficient completed trajectories have been collected for one training batch, the remaining in-flight rollouts are paused and their generated prefixes are retained. The same GPUs then switch to policy training. After the training states are offloaded and the updated model weights are synchronized to the inference engine, the paused requests resume generation from their retained prefixes.}
    \label{fig:partial_rollout}
\end{figure}

After evaluating the computational characteristics of the different RL stages and our infrastructure, we develop a co-located partial-rollout system based on the XTuner training engine and the LMDeploy inference engine. As illustrated in Figure~\ref{fig:partial_rollout}, the inference engine is continuously supplied with new requests during rollout generation to maintain high GPU utilization. Once the number of completed trajectories is sufficient to form a training batch, the remaining in-flight rollouts are paused at their current generation positions rather than aborted or discarded. Their generated prefixes and rollout metadata are retained, while the same GPU pool switches from inference to policy training. After the policy update, the training states are offloaded, the updated model parameters are synchronized to the inference engine, and the paused requests resume generation from their retained prefixes. Only completed trajectories are admitted into the current training batch.

This pause-and-resume mechanism avoids waiting for long-tail generations while preserving the computation already spent on unfinished responses. It also avoids the difficult producer--consumer balancing problem commonly encountered in fully asynchronous systems with disaggregated training and inference resources. However, because a resumed trajectory may contain segments generated before and after one or more policy updates, different tokens within the same trajectory can originate from different behavior-policy versions. We therefore record the behavior-policy version and generation-time log-probability for every sampled token.

For token \(y_{i,t}\) in trajectory \(i\), we define the importance-sampling ratio as
\begin{equation}
    \rho_{i,t}(\theta)
    =
    \frac{
        \pi_{\theta}(y_{i,t}\mid s_{i,t})
    }{
        \pi_{\mathrm{beh}(i,t)}(y_{i,t}\mid s_{i,t})
    },
    \label{eq:partial_rollout_is_ratio}
\end{equation}
where \(\pi_{\mathrm{beh}(i,t)}\) denotes the behavior-policy version that generated token \(y_{i,t}\), and \(\pi_\theta\) denotes the current learner policy. We explicitly bound trajectory staleness: a trajectory is discarded if its oldest retained segment was generated more than three policy updates before the current learner.

Following the clipped importance-weight formulation of \cite{chen2025minimax}, we truncate the importance ratio as
\begin{equation}
    \bar{\rho}_{i,t}(\theta)
    =
    \operatorname{clip}
    \left(
        \rho_{i,t}(\theta),
        1-\epsilon_{\mathrm{low}}^{\mathrm{IS}},
        1+\epsilon_{\mathrm{high}}^{\mathrm{IS}}
    \right).
    \label{eq:clipped_is_weight}
\end{equation}
The clipped ratio is subsequently used as a detached importance weight in the REINFORCE objective. Unlike PPO-style clipping, which clips the surrogate objective and may completely suppress gradients from tokens outside the trust region, clipping the importance weight bounds update variance while retaining a nonzero policy-gradient contribution from every unmasked token.

For MoE policies, training--inference inconsistency arises from both expert-routing differences and numerical discrepancies between the two execution engines. We employ Rollout Routing Replay (R3)~\cite{ma2025r3} to address the former: the expert selections made by LMDeploy during rollout are recorded and replayed by XTuner when evaluating the corresponding tokens. This ensures that rollout and training follow the same expert paths. Following Intern-S1-Pro~\cite{zou2026intern}, we additionally use an aligned mixed-precision configuration in which expert linear layers operate in FP8, the remaining layers use BF16, and numerically sensitive operations, including \texttt{apply\_rope}, RMSNorm, the MoE router, recurrent states in Gated DeltaNet~\cite{yang2025gateddeltanet}, and the language-model head, are computed in FP32.

After routing replay and operator-level alignment, a small number of tokens may still exhibit large probability discrepancies because of residual numerical differences. Inspired by KPop~\cite{ling26}, we detect these outliers using the bidirectional binary KL divergence. Let \(p_{i,t}^{\mathrm{train}}\) and \(p_{i,t}^{\mathrm{rollout}}\) denote the sampled-token probabilities produced by the training and rollout engines under matched model parameters and replayed routing decisions. We define
\begin{equation}
    D_{\mathrm{BKL}}(p\Vert q)
    =
    p\log\frac{p}{q}
    +
    (1-p)\log\frac{1-p}{1-q},
    \label{eq:binary_kl}
\end{equation}
and construct the token mask
\begin{equation}
    m_{i,t}^{\mathrm{BKL}}
    =
    \mathbb{I}
    \left[
        D_{\mathrm{BKL}}
        \left(
            p_{i,t}^{\mathrm{train}}
            \Vert
            p_{i,t}^{\mathrm{rollout}}
        \right)
        \leq \phi
    \right]
    \mathbb{I}
    \left[
        D_{\mathrm{BKL}}
        \left(
            p_{i,t}^{\mathrm{rollout}}
            \Vert
            p_{i,t}^{\mathrm{train}}
        \right)
        \leq \phi
    \right].
    \label{eq:bkl_token_mask}
\end{equation}
R3 removes discrete expert-routing mismatch, whereas the BKL mask filters the remaining token-level numerical outliers. The clipped importance weights in Equation~\eqref{eq:clipped_is_weight} and token masks in Equation~\eqref{eq:bkl_token_mask} are incorporated into the unified RL objective described in Section~\ref{sec:unified_reasoning_rl}.

\subsubsection{Adaptive Length Regularization for Efficient Reasoning}

Long-CoT reasoning models frequently exhibit overthinking on relatively simple problems, producing unnecessarily long reasoning trajectories even when the correct solution can be reached with substantially less computation~\cite{chen2024overthinking,pu2025thoughtterminator}.
Existing approaches improve reasoning efficiency through explicit length-aware rewards or constraints~\cite{kimi2025k15,aggarwal2025l1}, query-adaptive length penalties~\cite{xiang2025just}, or a separate length-control fine-tuning stage~\cite{luo2025o1pruner,zhao2025selfbraking}.
Although effective, reward-based approaches introduce auxiliary optimization objectives that may conflict with task rewards, whereas additional fine-tuning stages complicate the post-training pipeline and may disturb capabilities acquired during earlier RL stages.

We introduce an adaptive length regularization method that directly reweights the advantages of positive responses without adding an independent reward signal.
The method follows two principles.
First, we do not impose length regularization on negative responses.
Since an incorrect response may fail for many different reasons, penalizing its length can prematurely suppress potentially useful exploration and consequently degrade model performance.
Second, we activate length regularization only when the model achieves a sufficiently high pass rate on the corresponding query.
This design allows the model to freely explore difficult queries and encourages concise reasoning only after it has largely mastered them.

For each query \(q\), let \(\mathcal{G}_q=\{1,\ldots,G\}\) index a group of \(G\) sampled responses, and let \(\hat{A}_i\) denote the original advantage of response \(i\).
We define the set of positive responses as
\begin{equation}
    \mathcal{P}_q
    =
    \left\{
        i \in \mathcal{G}_q
        \mid
        \hat{A}_i > 0
    \right\}.
\end{equation}
The regularized advantage is given by
\begin{equation}
    \widetilde{A}_i
    =
    \begin{cases}
        \displaystyle
        \frac{
            \sum_{j \in \mathcal{P}_q} \hat{A}_j
        }{
            \sum_{j \in \mathcal{P}_q} w_j \hat{A}_j + \epsilon
        }
        w_i \hat{A}_i,
        &
        i \in \mathcal{P}_q,\;
        |\mathcal{P}_q| \geq \tau G,
        \\[12pt]
        \hat{A}_i,
        &
        \text{otherwise},
    \end{cases}
\end{equation}
where \(\tau\) controls the minimum fraction of positive responses required to activate length regularization.
The length-dependent weight \(w_i\) is defined as
\begin{equation}
    w_i
    =
    \alpha
    +
    (1-\alpha)
    \left(
        1-
        \frac{
            L_i-L_{\min}^{+}
        }{
            L_{\max}^{+}-L_{\min}^{+}+\epsilon
        }
    \right)^{\gamma},
    \qquad
    i \in \mathcal{P}_q,
\end{equation}
with
\begin{equation}
    L_{\min}^{+}
    =
    \min_{j \in \mathcal{P}_q} L_j,
    \qquad
    L_{\max}^{+}
    =
    \max_{j \in \mathcal{P}_q} L_j,
\end{equation}
where \(L_i\) is the reasoning length of response \(i\), \(\alpha\) specifies the minimum weight assigned to long responses, \(\gamma\) controls the shape of the length-dependent decay, and \(\epsilon\) ensures numerical stability.

Among positive responses, shorter solutions receive larger weights, whereas longer solutions are down-weighted while retaining positive advantages.
The normalization term approximately preserves the total positive advantage mass, thereby changing the relative preference among successful responses without substantially altering the overall optimization scale.
If the positive-response ratio is below \(\tau\), or if a response has a non-positive advantage, its advantage remains unchanged.
The method therefore adaptively transitions from exploration on difficult queries to efficiency optimization on queries that the model has already mastered.

\begin{figure}[t]
    \centering
    \includegraphics[width=0.85\linewidth]{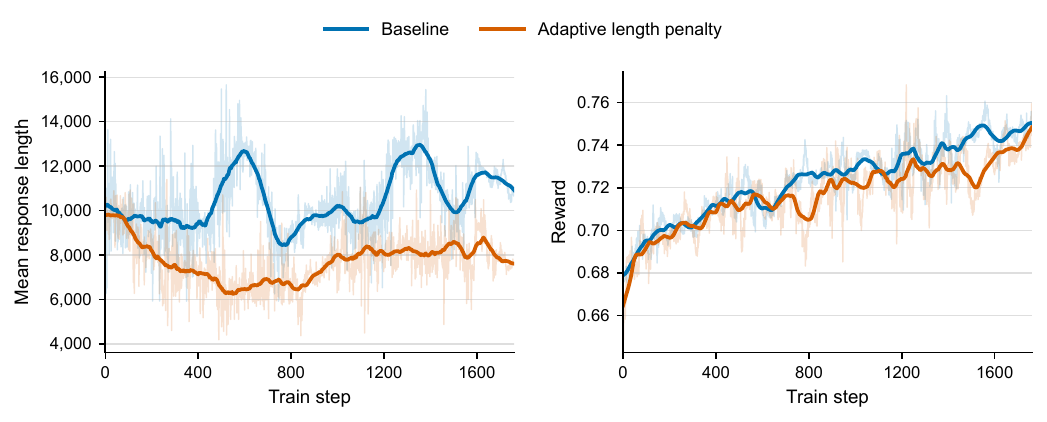}
    \caption{Comparison of training with and without adaptive length penalty, showing similar reward curves and shorter outputs with regularization.}
    \label{fig:adaptive_length_regularization}
\end{figure}

As shown in Figure~\ref{fig:adaptive_length_regularization}, we compare Intern-S2-Preview-35B trained with and without adaptive length regularization.
Both settings achieve comparable reward curves, while adaptive length regularization substantially reduces the average output length.
These results demonstrate that the proposed method improves reasoning efficiency without sacrificing model performance.

\subsubsection{Speculative Decoding for Faster RL Rollouts}

Although the co-located partial rollout system substantially improves GPU utilization, rollout generation remains one of the most time-consuming stages of RL training because long reasoning trajectories must still be generated autoregressively.
Speculative decoding provides a complementary approach for accelerating this process.
A lightweight draft model first proposes multiple candidate tokens, which are subsequently verified in parallel by the current policy model through an exact rejection-sampling procedure~\cite{leviathan2023fast,chen2023accelerating}.
Since this verification procedure preserves the sampling distribution of the policy model, speculative decoding accelerates rollout generation without introducing additional off-policy bias.
Recent studies have explored speculative decoding for RL rollouts through concurrency-aware online draft learning, continuously evolving draft models, tree-structured rollout caches, and system-level integration with RL infrastructure~\cite{zhang2025fastgrpo,chen2025respec,chang2026srt,iso2026accelerating}.

A central challenge in applying speculative decoding to RL is that the policy model evolves continuously during training.
A fixed draft model therefore becomes increasingly stale as the policy is updated, resulting in a growing mismatch between their output distributions and a progressive decline in the token acceptance rate.
To address this issue, we train the draft model online using trajectories generated by the latest policy.
At each RL iteration, the draft model is updated using the token distributions of the current policy on newly collected rollout states, while gradients are stopped through the policy model.
This online adaptation allows the draft model to continuously track the evolving rollout distribution throughout RL training.

We train the draft model using the hybrid LK Loss~\citep{samarin2026lk}, which combines the stable optimization behavior of forward KL divergence with the direct acceptance-rate optimization of total variation distance.
For the \(k\)-th draft position at a rollout state \(s_{t,k}\), we denote the target-policy and draft-model distributions as
\begin{equation}
    p_{t,k}(v)
    =
    \operatorname{sg}
    \left[
        \pi_{\theta}(v \mid s_{t,k})
    \right],
    \qquad
    q_{t,k}(v)
    =
    \pi_{\phi}^{\mathrm{draft}}(v \mid s_{t,k}),
\end{equation}
where both distributions are computed under the rollout sampling temperature and \(\operatorname{sg}[\cdot]\) denotes the stop-gradient operator.
The forward KL divergence and total variation distance are respectively defined as
\begin{equation}
    D_{\mathrm{KL}}
    \left(
        p_{t,k}
        \,\Vert\,
        q_{t,k}
    \right)
    =
    \sum_{v \in \mathcal{V}}
    p_{t,k}(v)
    \log
    \frac{
        p_{t,k}(v)
    }{
        q_{t,k}(v)
    },
\end{equation}
and
\begin{equation}
    D_{\mathrm{TV}}
    \left(
        p_{t,k},
        q_{t,k}
    \right)
    =
    \frac{1}{2}
    \sum_{v \in \mathcal{V}}
    \left|
        p_{t,k}(v)
        -
        q_{t,k}(v)
    \right|.
\end{equation}
Under lossless speculative sampling, the expected token acceptance probability is equal to the overlap between the target and draft distributions:
\begin{equation}
    \alpha_{t,k}
    =
    \sum_{v \in \mathcal{V}}
    \min
    \left(
        p_{t,k}(v),
        q_{t,k}(v)
    \right)
    =
    1
    -
    D_{\mathrm{TV}}
    \left(
        p_{t,k},
        q_{t,k}
    \right).
\end{equation}

Following the hybrid LK formulation, we combine the two divergence objectives as
\begin{equation}
    \mathcal{L}_{\mathrm{LK}}^{(t,k)}
    =
    \lambda_k
    D_{\mathrm{KL}}
    \left(
        p_{t,k}
        \,\Vert\,
        q_{t,k}
    \right)
    +
    \left(
        1-\lambda_k
    \right)
    D_{\mathrm{TV}}
    \left(
        p_{t,k},
        q_{t,k}
    \right).
\end{equation}
The mixing coefficient is adaptively determined by the acceptance rate:
\begin{equation}
    \lambda_k
    =
    \exp
    \left(
        -\eta\,
        \operatorname{sg}
        \left[
            \bar{\alpha}_k
        \right]
    \right),
    \qquad
    \eta > 0,
\end{equation}
where \(\bar{\alpha}_k\) is the acceptance rate for the \(k\)-th draft position aggregated over the sequence and batch dimensions.
The complete online draft-training objective is
\begin{equation}
    \mathcal{L}_{\mathrm{draft}}
    =
    \frac{1}{K}
    \sum_{k=1}^{K}
    \frac{1}{|\mathcal{T}_k|}
    \sum_{t \in \mathcal{T}_k}
    \mathcal{L}_{\mathrm{LK}}^{(t,k)},
\end{equation}
where \(K\) is the number of predicted draft positions and \(\mathcal{T}_k\) contains the valid training positions for the \(k\)-th draft step.
In our implementation, the draft model predicts tokens at \(K=4\) future positions, and we set the adaptive coefficient hyperparameter to \(\eta=3\).

When the draft model is poorly aligned with the current policy, the acceptance rate \(\bar{\alpha}_k\) is low and \(\lambda_k\) approaches one.
The objective is therefore dominated by the forward KL term, which provides smooth and well-scaled gradients for rapidly aligning the draft distribution with the evolving policy.
As the draft model becomes better aligned and the acceptance rate increases, \(\lambda_k\) decreases and the TV component receives a larger weight.
Since minimizing TV distance is equivalent to maximizing the distributional overlap, the objective gradually shifts from stable distribution matching to direct acceptance-rate optimization.

With online draft adaptation and the hybrid LK objective, the acceptance rate continues to improve as RL training progresses instead of degrading as the policy evolves.
In our large-scale training runs, speculative decoding ultimately delivers an approximately \(2\times\) speedup in rollout generation and a \(1.7\times\) end-to-end speedup for the overall RL training pipeline.
These results demonstrate that online draft learning provides a lossless and effective acceleration mechanism that complements our partial rollout system.

\subsubsection{Robust Multi-Task Optimization}
We perform RL for Intern-S2-Preview on mixtures of heterogeneous tasks, which differ in structure, solution diversity, and uncertainty of policy exploration.
This induces distinct entropy regimes under the same policy, making global or token-level entropy regulation inadequate for their heterogeneous exploration requirements~\cite{yang2025entropic,shen2025qwenlong,cheng2026reasoning,cui2025entropy}.
This heterogeneity further makes group-based policy optimization methods induce an entropy-dependent
bias, making advantage signals across prompt groups statistically non-comparable. 

We apply Group-level Entropy-Controlled Policy Optimization (GEPO)~\cite{cheng2026groupentropycontrolledpolicyoptimization}, which uses group-level entropy, estimated directly from existing grouped samples, as a diagnostic signal to identify and mitigate the optimization bias induced by entropy heterogeneity.
Specifically, GEPO attenuates positive advantages in low-entropy groups to prevent over-exploitation that would further amplify the entropy gap, while attenuating negative advantages in high-entropy groups to avoid prematurely suppressing exploration.
This scaling is asymmetric because low-entropy groups are more susceptible to aggressive intervention, which may trigger length collapse, and therefore require milder attenuation than high-entropy groups.

In GRPO and RLOO, given group responses $\{y_1, \ldots, y_K\}$ sampled from $\pi_\theta(\cdot|x)$ for each prompt $x$, we define group-level entropy as
\begin{equation}
    H_{\text{g}}(x) \;=\; -\frac{1}{K} \sum_{i=1}^{K} \sum_{t=1}^{T_i} \log \pi_\theta(y_{i,t} \mid y_{i,<t}, x).
    \label{eq:group_entropy_mc}
\end{equation}
Then, we shape the original advantage $\{A_i\}_{i=1}^K$ for each response as
\begin{equation}
    \hat{A}_i = \omega(g,A_i,\mathcal{H}_{\text{g}})A_i=
    \begin{cases}
        \alpha_{\text{low}}\cdot A_i  & \text{if } A_i > 0\ \text{and}\ \mathcal{H}_{\text{g}}(x) < \mathcal{H}_{\text{low}}^{(t)},\\
        \alpha_{\text{high}}\cdot A_i & \text{if } A_i < 0\ \text{and}\ \mathcal{H}_{\text{g}}(x) > \mathcal{H}_{\text{high}}^{(t)},\\
        A_i & \text{otherwise},
    \end{cases}
    \label{eq:adv_shaping}
\end{equation}
where $\alpha_{\text{high}}\in (0, 1) > \alpha_{\text{low}} \in (0, 1)$ are the scaling coefficients, and $\mathcal{H}_{\text{low}}^{(t)}$ and $\mathcal{H}_{\text{high}}^{(t)}$ denote the lower and upper entropy thresholds at training step $t$.

Instead of forcing heterogeneous tasks toward a shared entropy target, GEPO preserves task-dependent exploration regimes while rebalancing their effective contributions to policy updates.
It requires neither explicit task annotations nor additional rollouts and can be directly integrated into existing group-based policy optimization pipelines, providing a scalable mechanism for stable joint optimization over heterogeneous post-training tasks.

\subsubsection{Unified RL Objective and Training Configuration}
\label{sec:unified_reasoning_rl}

Our reasoning RL follows the leave-one-out REINFORCE formulation used in Intern-S1-Pro~\cite{zou2026intern}, augmented with the stabilization and efficiency techniques introduced above. For each query, we sample a group of \(G\) responses and obtain their sequence-level verifier rewards \(\{R_i\}_{i=1}^{G}\). Following the dynamic sampling strategy of DAPO~\cite{abs-2503-14476}, query groups whose rewards are all identical are filtered out online and replaced with newly sampled groups. For each retained group, the initial leave-one-out advantage is computed as

\begin{equation}
    A_i^{\mathrm{LOO}}
    =
    R_i
    -
    \frac{1}{G-1}
    \sum_{\substack{j=1\\j\neq i}}^{G}
    R_j.
    \label{eq:loo_advantage}
\end{equation}
We then apply the two advantage-shaping mechanisms described in the preceding sections. GEPO first adjusts the group-relative advantages according to group-level entropy to balance exploration across heterogeneous tasks. Adaptive length regularization is subsequently applied to the entropy-adjusted advantages, encouraging shorter successful reasoning trajectories only for query groups whose positive-response ratios exceed the activation threshold. The final advantage used for policy optimization can be summarized as
\begin{equation}
    \widetilde{A}_i
    =
    \mathcal{R}_{\mathrm{len}}
    \left(
        \mathcal{R}_{\mathrm{GEPO}}
        \left(
            A_i^{\mathrm{LOO}}
        \right)
    \right),
    \label{eq:final_shaped_advantage}
\end{equation}
where \(\mathcal{R}_{\mathrm{GEPO}}\) and \(\mathcal{R}_{\mathrm{len}}\) denote the entropy-control and adaptive length-regularization transformations defined above, respectively. This ordering ensures that the length-dependent weights act on the final entropy-adjusted advantages rather than modifying the verifier rewards.

Given a partial-rollout training buffer \(\mathcal{B}\), we optimize the policy using
\begin{equation}
    \mathcal{L}_{\mathrm{RL}}(\theta)
    =
    -
    \mathbb{E}_{
        \left(
            q,\{y_i\}_{i=1}^{G}
        \right)
        \sim\mathcal{B}
    }
    \left[
        \frac{1}{G}
        \sum_{i=1}^{G}
        \frac{1}{|y_i|}
        \sum_{t=1}^{|y_i|}
        m_{i,t}^{\mathrm{BKL}}\,
        \operatorname{sg}
        \left[
            \bar{\rho}_{i,t}(\theta)
        \right]
        \widetilde{A}_i
        \log
        \pi_{\theta}
        \left(
            y_{i,t}\mid s_{i,t}
        \right)
    \right],
    \label{eq:reasoning_rl_objective}
\end{equation}
where \(\bar{\rho}_{i,t}\) is the clipped token-level importance weight defined in Equation~\eqref{eq:clipped_is_weight}, \(m_{i,t}^{\mathrm{BKL}}\) is the numerical-consistency mask defined in Equation~\eqref{eq:bkl_token_mask}, and \(\operatorname{sg}[\cdot]\) denotes the stop-gradient operator. The sequence-level advantage \(\widetilde{A}_i\) is shared by all policy-generated tokens in response \(y_i\).

This objective combines three complementary stabilization mechanisms. The clipped importance weight corrects the policy mismatch introduced by pause-and-resume partial rollouts and repeated mini-batch updates. R3 aligns the expert-routing decisions of the rollout and training engines, while the BKL mask removes the remaining numerical outliers after routing and operator alignment. Meanwhile, GEPO and adaptive length regularization reshape the sequence-level advantages to balance task-dependent exploration and reasoning efficiency. Speculative decoding preserves the sampling distribution of the policy and therefore accelerates rollout generation without modifying Equation~\eqref{eq:reasoning_rl_objective}.

We use the Muon optimizer~\cite{jordan2024muon,liu2025muon} with a learning rate of \(1\times10^{-6}\) and a weight decay of \(0.01\). Each rollout batch contains \(8{,}192\) completed responses and is optimized through 8 mini-batch update steps. The maximum generation length is set to \(65{,}536\) tokens. Together, this training configuration and the stabilization mechanisms described above support efficient and stable reasoning RL over heterogeneous scientific and general-purpose tasks.

\subsection{Large-Scale Black- and White-Box Agentic RL}
\label{sec:agentic_rl}

Reasoning RL improves single-response and generation-oriented behavior, but scientific agents must also learn from interactive sessions that involve tools, files, external programs, and environment feedback.
We develop a unified agentic RL framework around a \emph{harness $\times$ task}
abstraction that decouples agent execution interfaces from task distributions.
A harness specifies how an agent is instantiated, driven, and observed, whereas
a task specifies the initial environment, executable objective, and
verifier-defined outcome. Their composition converts heterogeneous agent
executions into a common form of RL experience: an interactive rollout with an
explicit environment, an observable action--observation history, and automatic
outcome signals.

This formulation allows agentic training to scale along two complementary
axes. Along the harness axis, we support both white-box implementations whose
control loops can be directly orchestrated and black-box runtimes integrated
through their native CLI, SDK, or model API. Along the task axis, we cover
specialized coding and terminal environments as well as general-purpose tasks
produced by a self-evolving task-synthesis system. The resulting framework
broadens agent behaviors and task distributions while retaining a unified
rollout, verification, and training protocol.

Recent agentic RL systems have highlighted the importance of decoupling agent
execution from policy optimization and recovering trainable experience from
heterogeneous agent runtimes~\cite{luo2025agentlightning,xu2026polar,boyi2025adaptive}.
Intern-S2-Preview builds on a sequence of our studies on agent learning and
evaluation. Agent-FLAN examined data and optimization for effective agent
tuning; Lagent provided a modular framework for building language agents;
T-Eval and CIBench studied stepwise tool use and executable code-interpreter
behavior; MindSearch studied long-horizon information seeking and integration;
and SciExplore extended agent evaluation to realistic scientific navigation
and cross-source synthesis~\cite{chen2024agent,team2023lagent,chen2024teval,zhang2024cibench,liu2024mindsearch,tang2026sciexplore,zhao2024steve}.
More recent work investigates behavioral alignment through
process-continuation learning, the roles of next-chunk RL and SFT under
no-chain-of-thought supervision, and self-evolving task synthesis through skill
graphs and progressive validation~\cite{fang2026mindcompletion,fang2026mindcopilot,tang2026nextchunk,tang2026skill2task,zhao2024empowering,gao2025long}.
Together, these efforts motivate the \emph{harness $\times$ task} abstraction
developed here: a common system that can vary the agent runtime and task
distribution independently while retaining unified rollout, verification,
trace assembly, and RL optimization.

\subsubsection{Unified Agentic RL Infrastructure}
\label{sec:agentic_rl_infra}

\begin{figure}[t]
    \centering
    \includegraphics[width=\linewidth]{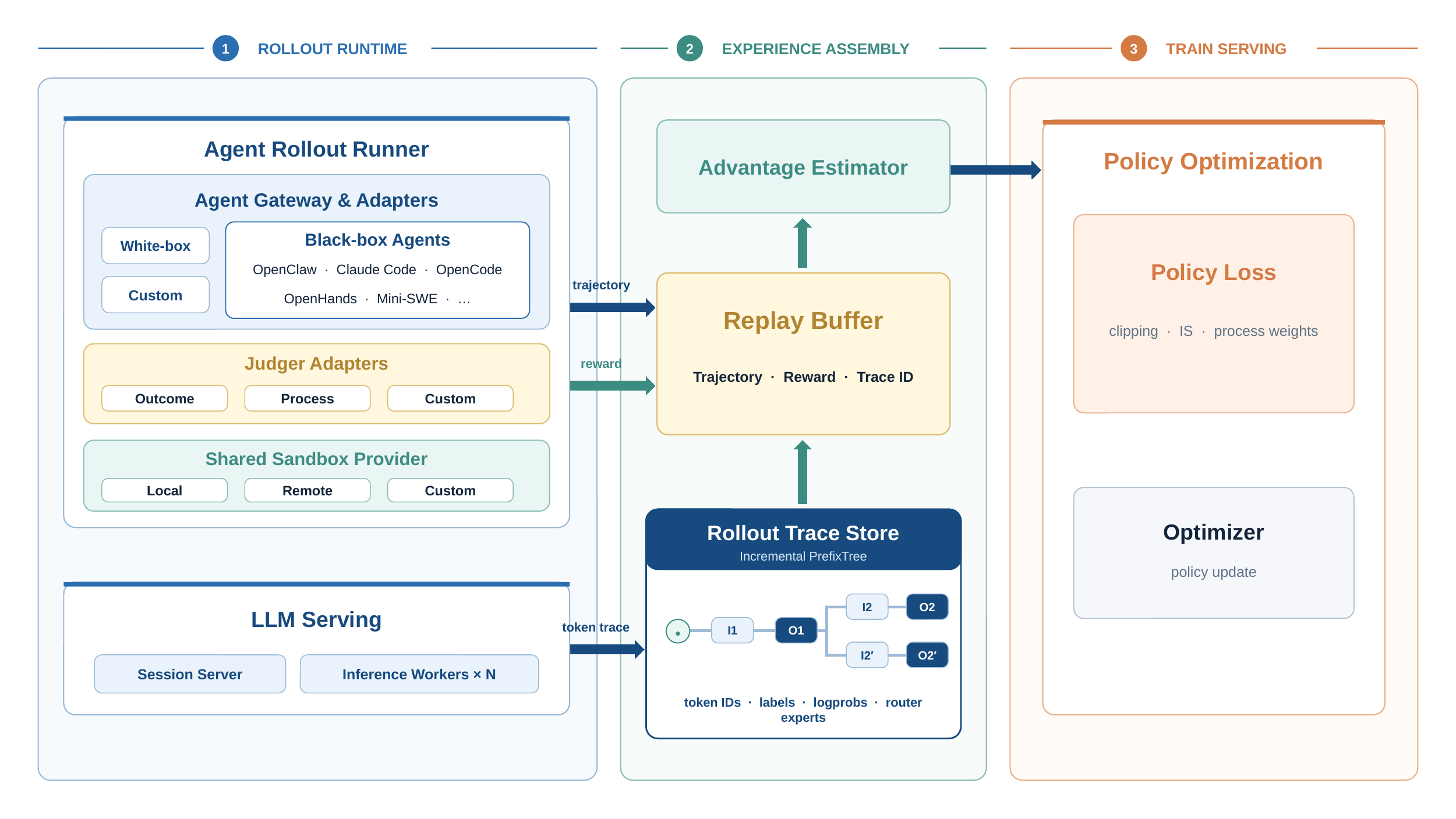}
    \caption{
        Overview of our agentic RL infrastructure. Heterogeneous white-box and
        black-box agents are unified by the Agent Gateway and execute against a
        shared sandbox and model-serving substrate. Semantic trajectories and
        verifier feedback are retained in the Replay Buffer, while the Rollout
        Trace Store preserves exact token-level evidence through a per-session
        incremental PrefixTree. Experience assembly aligns the two views for
        advantage estimation and policy optimization.
    }
    \label{fig:agentic_rl_infra}
\end{figure}

As illustrated in Figure~\ref{fig:agentic_rl_infra}, our infrastructure
consists of a unified rollout runtime and a trace-aware experience-assembly
layer. The former composes heterogeneous harnesses with executable tasks and
runs them under a common session contract; the latter joins semantic
trajectories and verifier feedback with the exact token-level evidence required
by policy optimization.

\noindent{\textbf{Unified execution runtime.}}
At rollout time, the Agent Rollout Runner receives a harness--task pair,
provisions its execution environment, and manages the interaction until normal
completion or a termination condition. The harness drives the agent, while
Judger Adapters perform outcome verification and process annotation against the
same session state and execution artifacts. A Shared Sandbox Provider abstracts
environment creation, command and tool execution, isolation, error handling,
and resource cleanup across local, remote, and custom backends. Agent control,
environment provisioning, and verification can therefore evolve independently
and be recombined across tasks without constructing a bespoke rollout pipeline
for every pairing.

\noindent{\textbf{Harness-agnostic agent integration.}}
Agent Gateway \& Adapters expose a stable interface over white-box, black-box,
and custom harness implementations. White-box loops can be orchestrated
directly, whereas black-box harnesses---including OpenClaw, Claude Code,
OpenCode, OpenHands, and Mini-SWE---retain their native messages, tool loops,
and control flow~\cite{wang2025openhands,yang2024sweagent}. The adapters
translate session lifecycle events, model calls,
and interaction artifacts without requiring the runtime to access or
reimplement a harness's internal agent logic. Adding a new harness therefore
requires a thin integration adapter rather than a new RL execution stack.

\noindent{\textbf{Client-transparent, training-aware model serving.}}
Different black-box harnesses expect different model protocols. Our LLM serving
layer accepts OpenAI Chat Completions, OpenAI Responses, and Anthropic Messages,
and supports both regular and streaming generation. Requests are normalized by
the gateway and forwarded to distributed inference workers, while streamed
text, reasoning, and tool-call events are relayed to the harness in their native
form. From the client's perspective, this remains an ordinary model service.
In parallel, the serving layer transparently captures training-only evidence,
including exact input and output token IDs, rollout log probabilities, and
token-wise MoE router experts, without exposing these extensions to the agent's
control logic.

This serving path implements a token-in--token-out (TITO) interface. For each
model call, the Session Server reuses the exact tokenized prefix already
recorded for the session, tokenizes only newly appended context, and sends the
resulting token IDs directly to the inference engine. The returned tokens and
policy statistics are captured from the same response stream delivered to the
agent. For sparse MoE models, Rollout Router Replay (R3) additionally records
the rollout-time expert choices for reuse during training. TITO therefore
preserves the sampled token sequence, while R3 preserves the conditional
computation path that produced it.

\noindent{\textbf{Trace-aware experience assembly.}}
The rollout runtime produces two complementary views of an interaction. The
Agent Runner and Judger Adapters produce the semantic experience---the
action--observation trajectory, outcome reward, process annotations, and session
metadata---which is retained in the Replay Buffer. LLM Serving produces the
model-execution evidence---token IDs, loss labels, behavior log probabilities,
and router experts---which is written to the Rollout Trace Store. Keeping these
views separate decouples environment-facing logic from model-specific training
representations while retaining a lossless path from an agent action to the
policy tokens that generated it.

The Trace Store organizes each session as an incremental PrefixTree. Each node
represents a newly appended context delta or assistant response and stores its
token IDs, labels, rollout log probabilities, and router experts. Longest-prefix
matching reuses the stable history of a session and appends only newly observed
segments. When a trajectory is selected for training, the store materializes
the corresponding root-to-leaf path. System instructions, user messages, and
tool observations are masked from the loss, while eligible policy-generated
segments retain their training labels.

Beyond incremental storage, the PrefixTree preserves the lineage and exact
boundaries of model calls across multi-turn and branching interactions. It thus
establishes a stable correspondence between a semantic agent action and its
rollout-time token span. Experience assembly joins Replay Buffer records and
token traces by session and segment, computes advantages, and exports
model-ready experiences to the common RL trainer. This correspondence also
provides the basis for the process-aware credit assignment described below.

\subsubsection{Agentic Task Construction}
\label{sec:agentic_rl_tasks}

We construct the task distribution from two complementary sources. For
specialized coding and terminal capabilities, we curate executable tasks from
publicly released collections containing both mined and synthetic instances.
For broader capabilities, a self-evolving task-synthesis system converts
community-contributed skills into general-purpose agentic tasks. Tasks from
both sources are normalized to a common contract consisting of an initialized
execution environment, a natural-language objective, and an automatic
verifier. Consequently, heterogeneous tasks can be composed with different
harnesses and trained within the same RL framework.

\noindent{\textbf{Coding and terminal tasks.}}
This branch draws from the public sources summarized in
Table~\ref{tab:agentic_task_sources}. Some collections mine real-world GitHub
issues, pull requests, and repository histories, whereas others procedurally
construct software-engineering environments or synthesize terminal and
workspace tasks. Together, they cover repository-level issue resolution,
debugging and testing, software setup, file and data manipulation, and broader
terminal-based problem solving.

Across sources, we map each instance into the common task contract by
materializing its base repository or container and required assets as the
initial environment, translating its issue statement or instruction into the
task objective, and retaining its tests or reward programs as the verifier.
This normalization preserves source-specific execution and reward semantics
while exposing a consistent interface to the rollout runtime. In particular,
these tasks remain grounded in live environments rather than being reduced to
static instruction--response pairs, so their rewards reflect program behavior,
repository state, and task-specific execution outcomes.

\begin{table*}[t]
\centering
\small
\caption{Public sources used to construct executable coding and terminal tasks.}
\label{tab:agentic_task_sources}
\begin{tabular}{llrr}
\toprule
Provider & Collection & \#Tasks & \#Environments \\
\midrule
\texttt{SWE-bench} & \texttt{SWE-smith}~\cite{yang2025swesmith} & 59,136 & 222 \\
\texttt{SWE-Gym} & \texttt{SWE-Gym}~\cite{pan2024swegym} & 2,438 & 2401 \\
\texttt{R2E-Gym} & \texttt{R2E-Gym-V1}~\cite{jain2025r2egym} & 7,480 & 8101 \\
\texttt{Nebius} & \texttt{SWE-rebench-V2}~\cite{badertdinov2026swerebenchv2} & 32,100 & 32075 \\
\texttt{AweAI-Team} & \texttt{Scale-SWE}~\cite{zhao2026scaleswe} & 20,200 & 19472 \\
\texttt{NVIDIA} & \texttt{Nemotron-Terminal-Synthetic-Tasks}~\cite{pi2026terminal} & 80,000 & 8 \\
\texttt{RUC-AIBOX} & \texttt{ClawGym-Task}~\cite{bai2026clawgym} & 13,500 & 1 \\
\bottomrule
\end{tabular}
\end{table*}

\begin{figure}[t]
    \centering
    \includegraphics[width=\linewidth]{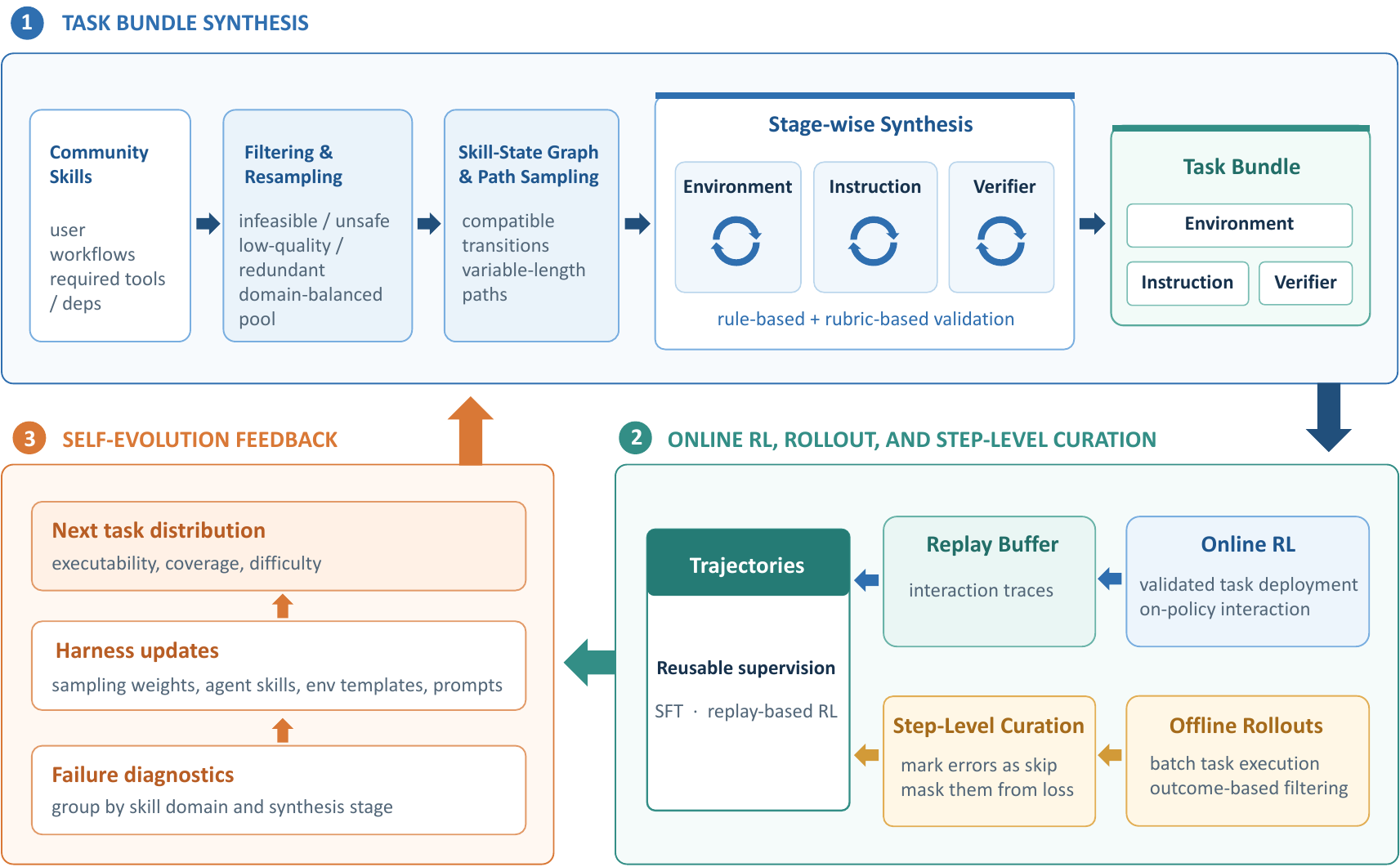}
    \caption{
        Self-evolving construction of general agentic tasks. Community skills
        are filtered and composed through compatible skill-state paths before
        stage-wise synthesis produces validated task bundles. Online and
        offline rollouts yield curated reusable trajectories, while execution
        failures update sampling and synthesis components to generate the next
        task distribution.
    }
    \label{fig:skill2task_self_evolve}
\end{figure}

\noindent{\textbf{Self-evolving general agentic tasks.}}
For broader agentic coverage, we build a self-evolving task-synthesis system
organized as the closed data loop illustrated in
Figure~\ref{fig:skill2task_self_evolve}. We seed the system with
community-contributed skills collected from multiple sources. These skills
describe concrete user workflows and their required tools and dependencies,
providing a natural basis for executable task synthesis. We first filter
infeasible, unsafe, low-quality, or redundant candidates, including workflows
involving unavailable authentication, external transactions, or toxic content,
and then perform domain-balanced resampling to prevent frequently occurring
skill domains from dominating the synthesis distribution.

Individual skills often describe localized workflows, whereas a general agent
must coordinate multiple capabilities over longer horizons. We therefore build
a skill-state graph whose nodes denote observable environment states and whose
edges denote state-transforming capabilities extracted from skills. Skills are
composed only when their input and output states are compatible. Sampling paths
of different lengths from this graph produces capability sequences with varied
horizons and complexity.

Each sampled path is converted into an executable task bundle through a
progressive pipeline that constructs the environment, task, and verifier in
sequence. Every stage is paired with an executable validator. Rule-based checks
verify structural correctness, dependency resolution, and executability, while
rubric-based checks evaluate semantic quality and cross-stage consistency.
Failures trigger stage-local repair or regeneration, and only validated
artifacts are passed downstream. This localizes synthesis errors and reduces
inconsistencies among the environment, task specification, and verifier.

Validated tasks are deployed in online RL and are also rolled out to construct
reusable offline training data. Candidate trajectories passing outcome-based
filtering undergo step-level curation. Each interaction step is annotated by
behavior type, covering normal progress as well as tool-use errors, repetitive
failed attempts, invalid recovery, premature termination, protocol violations,
unsupported assumptions, and hallucinated observations. Erroneous steps remain
in the interaction context but can be marked as \texttt{skip} and excluded from
the imitation loss, while the remaining responses serve as optimization
targets. This selective masking avoids imitating flawed intermediate behavior
without destroying the causal context of later actions.

Downstream execution further exposes systematic weaknesses in the synthesized
task distribution. We aggregate execution failures and verifier feedback by
skill domain and synthesis stage, and use these statistics to update skill
sampling weights as well as synthesis skills, environment templates, and
stage-specific prompts. The revised system then resamples capability paths and
generates the next task distribution, progressively improving executability,
coverage, and difficulty from observed agent behavior.

\subsubsection{Robust Agentic RL Training}
\label{sec:agentic_rl_training}

Optimizing on agentic experience introduces challenges beyond single-turn
response optimization. Verifiers score complete, tool-interleaved sessions,
while only selected policy-generated segments are trainable. Moreover,
executable environments expose the reward channel to solution leakage, test
manipulation, and other forms of reward hacking. We first align terminal
outcomes with the trainable segments of a session, then refine this signal with
process-aware advantage control, and finally harden the verifier against reward
leakage. We conclude by examining the resulting optimization behavior across
task families and agent harnesses.

\noindent{\textbf{Session-aware outcome credit.}}
An agentic session may contain multiple assistant responses separated by
system instructions, user messages, tool calls, and environment observations.
Nevertheless, these segments jointly solve one task and receive one final
outcome reward. Within each rollout group for the same task, we compute a
group-relative advantage $A_i$ from the complete-session reward of rollout
$i$, following the group-relative optimization paradigm~\cite{abs-2402-03300}.
The same session advantage is assigned to all eligible policy-generated
segments in its trace, rather than treating each model call as an independently
rewarded episode. Token-level labels exclude non-policy context from the loss,
while the PrefixTree preserves the exact boundaries of the trainable response
segments.

\noindent{\textbf{Process-aware advantage control.}}
Outcome-only credit can reinforce undesirable intermediate behavior when a
session eventually succeeds despite malformed outputs, invalid or repeated
tool calls, unnecessary recovery attempts, or abnormal termination. We keep
outcome evaluation and process feedback separate, reflecting the distinction
between outcome and process supervision~\cite{LightmanKBEBLLS24}. The outcome verifier
determines whether the task is solved, while a process annotator attaches an
\texttt{adv\_penalty} to the specific assistant message exhibiting a
deterministic process error. These annotations do not change the session reward
or the token labels.

\begin{figure}[t]
    \centering
    \includegraphics[width=0.96\linewidth]{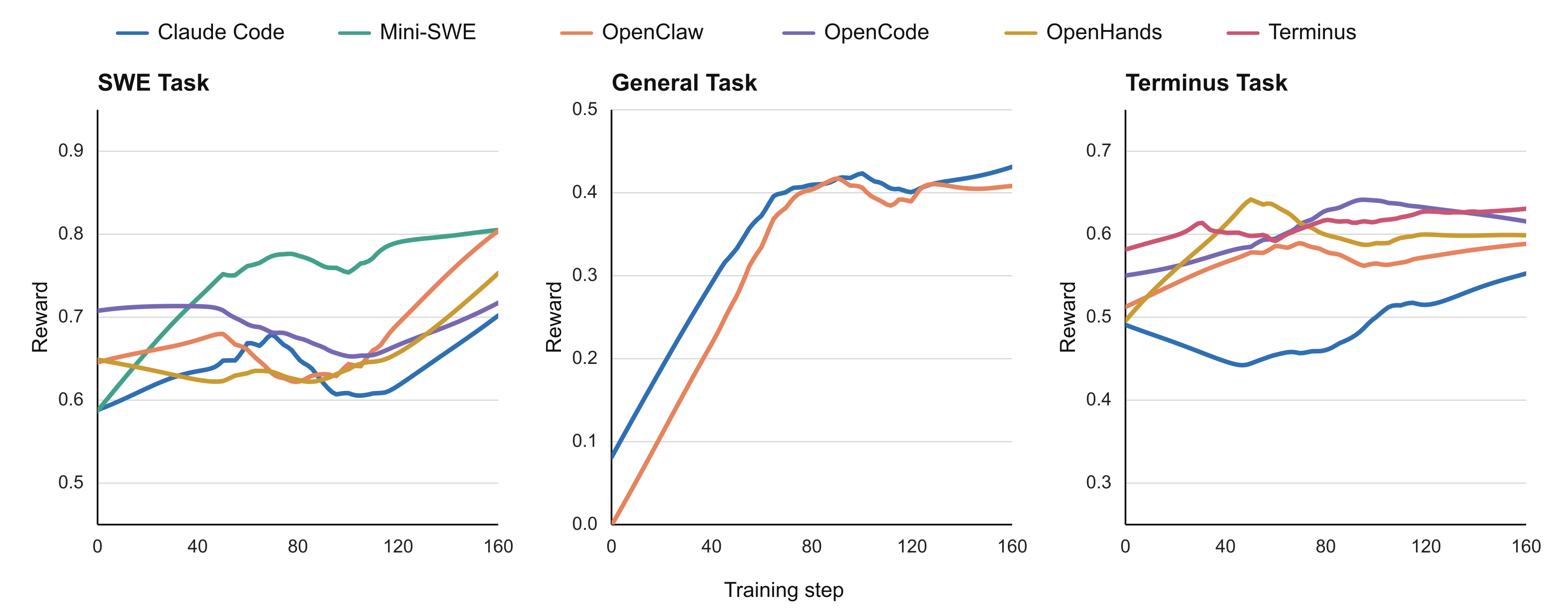}
    \caption{
        Reward trajectories across SWE, general-purpose, and terminal tasks
        under multiple agent harnesses. Each panel presents a representative
        example over 160 optimization steps. Curves are locally smoothed to
        highlight the overall optimization trends.
    }
    \label{fig:agentic_rl_reward_curves}
\end{figure}

The PrefixTree maps each annotated message to its exact trainable token span.
Let $w_{i,k}\in[-1,1]$ denote the process weight for segment $k$ in session $i$. For a
trainable token $t$ in that segment, we use
\begin{equation}
    \widetilde{A}_{i,k,t}
    =
    \begin{cases}
        w_{i,k} A_i, & A_i > 0,\\
        A_i, & A_i \leq 0,
    \end{cases}
\end{equation}
and set the advantage to zero for non-trainable tokens. Process weights can
suppress or reverse positive credit for parse and format errors, invalid tool
names or arguments, repeated or failed tool calls, and context-, turn-, or
session-limit termination. They are applied only to positive advantages, so
the negative learning signal of failed trajectories is preserved. In short,
the outcome reward determines whether the task is solved, whereas the process
weight determines whether an intermediate behavior should receive positive
credit.

\noindent{\textbf{Verifier integrity and leakage prevention.}}
For executable coding tasks, we isolate agent-visible execution artifacts from
grading-only information. Gold patches, held-out tests, and exact scoring test
identifiers are excluded from the rollout workspace and made available only to
the grading infrastructure after agent execution. Repository histories are
sanitized into a single baseline commit and remote references are removed,
preventing an agent from recovering target fixes through Git metadata. Task
identifiers that directly reveal upstream issues are likewise omitted from
agent-facing instructions when necessary.

During evaluation, canonical test files are restored or overlaid after the
agent has stopped, and gold test patches are applied on top of the agent's
source changes. Modifications to agent-visible tests therefore cannot directly
alter the scoring procedure. We use conservative all-correct semantics for
software-engineering tasks, requiring both target-fix and regression checks to
pass; for tasks with a canonical expected test-state map, the observed outcomes
must match it exactly. Missing grading artifacts, execution failures, and
unparseable verifier outputs are tracked separately from genuine task failure,
preventing infrastructure errors from being interpreted as successful policy
behavior. Together, these measures keep the reward tied to genuine changes in
the executable task state rather than access to hidden solutions or corruption
of the verifier.

\noindent{\textbf{Optimization across tasks and harnesses.}}
The same training and experience-assembly path is used across software
engineering, general-purpose, and terminal tasks, without introducing a
task-specific RL pipeline for each harness. Figure~\ref{fig:agentic_rl_reward_curves}
summarizes representative reward trajectories. General-purpose tasks improve
rapidly and then stabilize for both Claude Code and OpenClaw. The longer-horizon
SWE and terminal tasks exhibit more harness-dependent transients, but their
displayed trajectories improve or recover over the optimization window. These
different dynamics are expected: a harness determines the interaction policy
and context construction, while the task determines the environment and reward
semantics. The shared upward trend is therefore evidence that the common trace,
credit-assignment, and optimization stack remains effective across distinct
harness--task compositions, rather than evidence that their absolute reward
scales are directly comparable.

\subsection{On-Policy Distillation}

The final stage merges the complementary strengths of the separately optimized reasoning and agentic policies into the released unified model.
Although mixed reinforcement learning enables a single policy to acquire broad capabilities, jointly optimizing highly heterogeneous reasoning and agentic tasks can introduce optimization conflicts and prevent the model from fully exploiting domain-specific training signals. We therefore separately train two expert policies from the same SFT checkpoint: a reasoning expert obtained through mixed reasoning RL and an agentic expert obtained through large-scale black-box and white-box agentic RL. We then employ on-policy distillation (OPD) to consolidate their complementary capabilities into a single unified model. 
Unlike prior multi-teacher approaches that train a large number of fine-grained domain experts~\cite{ma2026mopd,nvidia2026nemotron3ultra,zhao2024we,zhao2024hierarchical}, we organize specialization around two broad capability domains. Our preliminary evaluation suggests that independently training teachers for many fine-grained domains incurs substantial RL and infrastructure costs, while providing limited additional benefit for our setting. The two-expert design achieves a favorable balance between specialization quality, training cost, and distillation complexity.

A key challenge in OPD is the distribution mismatch between the student and its teachers. Since teachers are evaluated on prefixes sampled by the student, a large policy discrepancy can cause student trajectories to fall outside the reliable support of the teacher, resulting in noisy or uninformative supervision. In our setting, both expert policies are derived from the same SFT checkpoint and therefore retain broadly compatible generation behaviors. To further reduce the initial discrepancy, we follow the warmup strategy introduced in Nemotron 3 Ultra~\cite{nvidia2026nemotron3ultra}. Specifically, we use the reasoning and agentic experts to generate high-quality trajectories and perform a lightweight SFT warmup on the original SFT model. The resulting checkpoint serves as the initial student for OPD. 
This warmup exposes the student to the characteristic reasoning patterns and interaction behaviors of both teachers, increasing the overlap between student-generated trajectories and teacher-supported distributions before OPD optimization begins.

During OPD, each query is assigned to either the reasoning or agentic domain, and the corresponding expert is selected as its teacher. Let \(d \in \{\mathrm{rea},\mathrm{agt}\}\) denote the domain, \(\mathcal{D}_d\) its prompt distribution, and \(\pi_{T_d}\) the corresponding teacher policy. Given a query \(q\), the student policy \(\pi_\theta\) first generates an on-policy trajectory \(y\), and the selected teacher evaluates the student-generated prefixes \(s_t=(q,y_{<t})\). Following Nemotron 3 Ultra~\cite{nvidia2026nemotron3ultra}, the fully on-policy objective maximizes the negative reverse KL divergence:
\begin{equation}
    \mathcal{J}_{\mathrm{OPD}}(\theta)
    =
    \sum_{d \in \{\mathrm{rea},\mathrm{agt}\}}
    \lambda_d
    \mathbb{E}_{
        q\sim\mathcal{D}_d,\,
        y\sim\pi_\theta(\cdot\mid q)
    }
    \left[
        \sum_{t=1}^{H}
        \left(
            \log\pi_{T_d}(y_t\mid s_t)
            -
            \log\pi_\theta(y_t\mid s_t)
        \right)
    \right],
\end{equation}
where \(\lambda_d\) controls the sampling or loss weight of each domain and \(H\) denotes the trajectory length. Equivalently, this objective minimizes
\begin{equation}
    D_{\mathrm{KL}}
    \left(
        \pi_\theta(\cdot\mid s_t)
        \,\Vert\,
        \pi_{T_d}(\cdot\mid s_t)
    \right)
\end{equation}
on states induced by the student itself. In contrast to RL objectives based on sparse trajectory-level rewards, OPD provides dense token-level supervision from the corresponding expert throughout the generated trajectory.

Existing OPD systems may transfer the complete teacher distribution or its top-\(k\) approximation at every token position~\cite{deepseekai2026deepseekv4,ma2026mopd}. However, the communication and storage costs of these approaches become substantial for our maximum sequence length of 256K tokens. Transmitting either full-vocabulary logits or even the top-64 teacher logits for every position introduces a large communication payload between teacher-scoring workers and learner workers. Since our teachers share the same SFT origin and the warmup stage further reduces their policy discrepancy with the student, we find that transmitting only the teacher log-probability of each sampled token is sufficient for stable distillation. This reduces the teacher payload from \(O(HV)\) or \(O(Hk)\) to \(O(H)\), where \(V\) is the vocabulary size and \(k\) is the number of retained teacher logits.

To support policy updates over trajectories produced by our partial-rollout infrastructure, we use the same clipped importance-weighted REINFORCE formulation as in our reasoning RL. The only difference lies in the construction of the advantage. For each sampled token \(y_{i,t}\), we define the sampled-token distillation advantage as
\begin{equation}
    \widehat{A}_{i,t}^{\mathrm{OPD}}
    =
    \operatorname{sg}
    \left[
        \log
        \pi_{T_d}
        \left(
            y_{i,t}\mid s_{i,t}
        \right)
        -
        \log
        \pi_{\mathrm{prox}}
        \left(
            y_{i,t}\mid s_{i,t}
        \right)
    \right],
    \label{eq:opd_advantage}
\end{equation}
where \(\pi_{T_d}\) is the teacher assigned to domain \(d\), \(\pi_{\mathrm{prox}}\) is the frozen proximal student policy used to construct the distillation signal, and \(\operatorname{sg}[\cdot]\) denotes the stop-gradient operator. The advantage is positive when the teacher assigns a higher probability to the sampled token than the proximal student and negative otherwise, thereby increasing or decreasing the probability of the sampled action accordingly.

As in Equation~\eqref{eq:partial_rollout_is_ratio}, the token-level importance-sampling ratio is defined directly between the current learner policy and the behavior-policy version that generated the token:
\begin{equation}
    \rho_{i,t}^{\mathrm{OPD}}(\theta)
    =
    \frac{
        \pi_{\theta}
        \left(
            y_{i,t}\mid s_{i,t}
        \right)
    }{
        \pi_{\mathrm{beh}(i,t)}
        \left(
            y_{i,t}\mid s_{i,t}
        \right)
    }.
\end{equation}
We clip this ratio using the same interval as reasoning RL:
\begin{equation}
    \bar{\rho}_{i,t}^{\mathrm{OPD}}(\theta)
    =
    \operatorname{clip}
    \left(
        \rho_{i,t}^{\mathrm{OPD}}(\theta),
        1-\epsilon_{\mathrm{low}}^{\mathrm{IS}},
        1+\epsilon_{\mathrm{high}}^{\mathrm{IS}}
    \right).
    \label{eq:opd_clipped_is_weight}
\end{equation}

Let \(\mathcal{B}_d\) denote a partial-rollout batch assigned to domain \(d\), \(N_d\) the number of trajectories from this domain, and \(\mathcal{T}_i\) the set of policy-generated token positions in trajectory \(i\). We optimize the student using
\begin{equation}
    \mathcal{L}_{\mathrm{OPD}}(\theta)
    =
    -
    \sum_{d\in\{\mathrm{rea},\mathrm{agt}\}}
    \lambda_d
    \mathbb{E}_{\mathcal{B}_d}
    \left[
        \frac{1}{N_d}
        \sum_{i=1}^{N_d}
        \frac{1}{|\mathcal{T}_i|}
        \sum_{t\in\mathcal{T}_i}
        m_{i,t}^{\mathrm{BKL}}\,
        \operatorname{sg}
        \left[
            \bar{\rho}_{i,t}^{\mathrm{OPD}}(\theta)
        \right]
        \widehat{A}_{i,t}^{\mathrm{OPD}}
        \log
        \pi_{\theta}
        \left(
            y_{i,t}\mid s_{i,t}
        \right)
    \right],
    \label{eq:opd_objective}
\end{equation}
where \(\lambda_d\) controls the contribution of each teacher domain and \(m_{i,t}^{\mathrm{BKL}}\) is the same numerical-consistency mask used in reasoning RL. Non-policy tokens, including prompts, environment observations, tool outputs, and padding tokens, are excluded through \(\mathcal{T}_i\).

Equation~\eqref{eq:opd_objective} has the same optimization form as the unified reasoning RL objective in Equation~\eqref{eq:reasoning_rl_objective}. In both cases, the current policy is optimized using a detached and clipped behavior-to-current importance weight, together with R3 routing alignment and BKL-based numerical masking. The only distinction is the advantage: reasoning RL uses a sequence-level advantage derived from verifier rewards, GEPO, and adaptive length regularization, whereas OPD uses the token-level teacher--student log-probability difference in Equation~\eqref{eq:opd_advantage}.

Through shared initialization, teacher-trajectory warmup, and sampled-token on-policy supervision, our approach avoids the complexity and communication overhead of full-vocabulary or top-\(k\) distillation while maintaining stable optimization over long trajectories. The resulting student consolidates the scientific reasoning capabilities of the mixed-reasoning policy and the long-horizon interaction capabilities of the agentic policy into the unified Intern-S2-Preview model.

%% file: sections/5.evaluation.tex
\newcommand{\open}[1]{\underline{#1}}
\newcommand{\best}[1]{\textbf{#1}}

\setlength{\aboverulesep}{0pt}
\setlength{\belowrulesep}{0pt}
\renewcommand{\arraystretch}{1.25} %

\definecolor{internblue}{RGB}{220,235,247}
\definecolor{subgray}{RGB}{245,245,245}
\newcolumntype{L}[1]{>{\raggedright\arraybackslash}p{#1}}
\newcolumntype{D}[1]{>{\raggedright\arraybackslash}p{#1}}
\newcolumntype{C}[1]{>{\centering\arraybackslash}m{#1}}
\newcolumntype{B}[1]{>{\columncolor{internblue}\centering\arraybackslash}m{#1}}

\newcommand{\SubTableTitle}[1]{%
    \multicolumn{6}{l}{\hspace{-2pt}\textbf{#1}} \vspace{4pt} \\
}   
\newcommand{\modelhead}[3]{%
  \makecell[c]{\small\textbf{#1}\\\small\textbf{#2}}%
}

\section{Evaluation}

We conduct extensive experiments to evaluate Intern-S2-Preview-397B across a wide range of benchmarks from two perspectives: scientific tasks and general-purpose tasks, covering both text-only and multimodal settings. In this section, we first introduce the evaluation setup, followed by a brief description of the benchmarks employed. We then compare the performance of Intern-S2-Preview-397B with other state-of-the-art models.
 
\subsection{Benchmarks}

\subsubsection{Scientific benchmarks}

\textbf{Biology-Instructions} \cite{he2025biology} is a multi-omics benchmark that evaluates the sequence understanding capabilities of models across diverse biological scales. It integrates biological sequence-based prediction tasks with advanced reasoning requirements, challenging models to interpret complex genomic, transcriptomic, and proteomic data.

\textbf{Mol-Instructions} \cite{fang2024molinstructionslargescalebiomolecularinstruction} is designed to bridge the gap in specialized LLM training through three primary categories: molecule-oriented, protein-oriented, and biomolecular text-oriented tasks. It includes a vast collection of instruction-following pairs that facilitate the model's proficiency in handling complex biomolecular structures and functional descriptions.

\textbf{MolecularIQ} \cite{bartmann2026moleculariqcharacterizingchemicalreasoning} evaluates the ability of language models to reason faithfully over molecular graphs represented as SMILES. It contains 5,111 symbolically verifiable questions involving 849 structurally held-out molecules and organizes them into counting, indexing, and constrained-generation task families. Its sampling procedure balances reasoning depth, multitask load, molecular complexity, and answer distributions, enabling fine-grained localization of failures in molecular-structure reasoning.

\textbf{SciReasoner} \cite{wang2025scireasonerlayingscientificreasoning} evaluates scientific reasoning across diverse disciplines, including physics, chemistry, and medicine, 9 domains in total and 149 concrete tasks. It consists of a unified suite of ten sub-benchmarks with varying question formats such as multiple-choice, fill-in-the-blank, and protocol-based procedural questions, designed to assess both knowledge retrieval and complex deductive reasoning. 

\textbf{TOMG-Bench} \cite{li2024tomgbenchevaluatingllmstextbased} evaluates open-domain, natural-language-guided molecule generation. It comprises three major tasks---molecule editing, molecular-property optimization, and customized molecule generation---each divided into three subtasks with 5,000 test samples per subtask. An automated evaluation framework measures whether generated molecules are valid, satisfy the requested structural or property constraints, and retain appropriate similarity or novelty.

\textbf{MP20} is a conditional crystal structure generation benchmark for materials with at most 20 atoms per unit cell. The task requires predicting precise atomic coordinates and lattice parameters from chemical compositions under physical constraints such as periodicity and symmetry. It contains 27,136 training and 9,046 test samples with ground-truth structural targets provided without chain-of-thought reasoning annotations.

\textbf{ProteinBinder-9} is a focused benchmark for evaluating de novo protein binder design across nine biologically relevant protein targets. The benchmark is designed in the spirit of the protein–protein interaction design setting introduced by ODesign, an all-atom generative world model for biomolecular interaction design~\cite{zhang2025odesign}. For each target, the task is to generate protein binders that satisfy a predefined binding interface and pass a multi-stage structural and physicochemical evaluation pipeline. Candidate backbones and sequences are generated and evaluated using complementary computational models, including structure-generation methods such as RFdiffusion~\cite{watson2023novo} and biomolecular complex prediction with AlphaFold 3~\cite{abramson2024accurate}. The resulting candidates are further assessed using interface-confidence, binding-energy, and molecular-contact criteria. ProteinBinder-9 covers diverse target proteins and interface geometries, thereby testing whether a design system can generalize beyond a single protein family or structural context. We report both the number and fraction of candidates that pass the complete evaluation procedure for each target, together with aggregate results across all nine targets. ProteinBinder-9 is intended to provide a reproducible testbed for evaluating end-to-end protein binder design systems, rather than isolated sequence generation or structure prediction performance.

\textbf{XLRS-Bench} \cite{Wang_2025_CVPR} focuses on extremely large, ultra-high-resolution remote sensing (RS) imagery; this benchmark defines 16 sub-tasks to evaluate 6 types of perceptual and 4 types of reasoning abilities. It challenges MLLMs to process complex semantic relationships and facilitate real-world decision-making in high-resolution geospatial scenarios.

\textbf{MicroVQA} \cite{burgess2025microvqamultimodalreasoningbenchmark} focuses on microscopy-based research and consists of 1,042 expert-curated multiple-choice questions across diverse imaging modalities. The benchmark assesses three critical reasoning capabilities within biological workflows: expert image understanding, hypothesis generation, and experimental proposal.

\textbf{SFE} \cite{zhou2025scientistsexamprobingcognitive} is an expert-level benchmark comprising 830 verified visual question answering (VQA) pairs across 66 multimodal tasks. Spanning five high-value scientific disciplines, the dataset utilizes authentic raw scientific data formats to probe the cognitive abilities of models in perception, understanding, and advanced reasoning.

\textbf{ObsCrisis-Bench} \cite{obscrisis-bench} evaluates multimodal reasoning about extreme-weather and geophysical crises from multispectral satellite observations and optional weather-station measurements. Its official dataset card reports 4,202 visual-question-answering samples covering 127 events, eight disaster categories, and 61 countries. The tasks span early warning, event-type and timing prediction, impact assessment, and post-event recovery analysis across multiple observation timesteps.

\textbf{SciCode} \cite{tian2024scicode} evaluates the ability of language models to write executable code for realistic scientific research problems rather than conventional algorithmic exercises. It contains 80 main problems decomposed into 338 subproblems across 16 scientific subfields, including physics, mathematics, materials science, biology, and chemistry. Each problem combines scientific knowledge, reasoning, and code synthesis and is accompanied by optional background material, scientist-written reference solutions, and executable tests.

\textbf{SGI-Bench} \cite{xu2025probingscientificgeneralintelligence} evaluates Scientific General Intelligence across the complete inquiry cycle defined by the Practical Inquiry Model: deliberation, conception, action, and perception. It contains 1,263 expert-curated samples spanning ten scientific domains and 75 research directions, organized into scientific deep research, idea generation, dry and wet experiments, and multimodal experimental reasoning. Task-specific multidimensional metrics assess evidence synthesis, novelty and feasibility, computational correctness, protocol fidelity, and interpretation of experimental results.

\textbf{ResearchClawBench} \cite{xu2026researchclawbenchbenchmarkendtoendautonomous} evaluates whether autonomous agents can conduct end-to-end scientific research from raw data and related literature to a publication-style research report. It contains 40 tasks derived from real papers across ten scientific domains, while withholding the target paper during evaluation. Expert-authored, multimodal rubrics measure whether agents reproduce the original experimental protocols, evidence chains, analyses, and scientific conclusions while leaving room for findings beyond the source paper. We evaluated ResearchClawBench on ResearchHarness v0.0.49.

\subsubsection{General benchmarks}

\textbf{MMLU-Pro} \cite{wang2024mmluprorobustchallengingmultitask} enhances the original MMLU by increasing the number of choices and focusing on more challenging, reasoning-intensive questions. It covers a broad range of subjects, requiring models to demonstrate deeper multi-task language understanding and more robust problem-solving skills.

\textbf{SimpleQA-Verified} \cite{haas2025simpleqaverifiedreliablefactuality} evaluates short-form factuality and parametric knowledge without access to retrieval tools. It consists of 1,000 human-verified prompts derived from SimpleQA through deduplication, topic balancing, source reconciliation, ambiguity removal, and adversarial difficulty filtering. Its revised autorater distinguishes correct, incorrect, and not-attempted responses while handling numeric tolerances, hedging, and answer-format variation more reliably.

\textbf{AdvancedIF} \cite{he2025advancedifrubricbasedbenchmarkingreinforcement} evaluates advanced instruction following under complex single-turn instructions, multi-turn carried context, and system-prompt steerability. It contains 1,645 human-written prompts paired with expert-authored rubrics of up to 20 independently verifiable criteria. A response succeeds only when it satisfies all applicable criteria, making the benchmark sensitive to subtle omissions and conflicts across user, conversational, and system-level instructions.

\textbf{HMMT-2026} \cite{dekoninck2026benchmarksmatharenaevaluationplatform} evaluates advanced mathematical reasoning using 33 problems from the February 2026 Harvard--MIT Mathematics Tournament released through MathArena. The problems cover algebra, combinatorics, geometry, and number theory and were converted to LaTeX and manually verified together with their reference answers. Because the problems were evaluated soon after the competition, the benchmark also serves as a relatively fresh test of competition-level mathematical problem solving.

\textbf{MMMU-Pro} \cite{yue2025mmmuprorobustmultidisciplinemultimodal} is a robust extension of the MMMU benchmark. MMMU-Pro introduces more challenging multidisciplinary multimodal tasks. It emphasizes expert-level understanding and complex reasoning across a wide array of professional domains, utilizing high-resolution images and specialized knowledge.

\textbf{ChartQAPro} \cite{masry-etal-2025-chartqapro} evaluates visual perception and complex reasoning over realistic charts. The published benchmark contains 1,341 charts collected from 99 diverse sources and 1,948 questions covering mathematical and visual reasoning, conversational queries, fact checking, multiple-choice questions, and hypothetical scenarios. It includes heterogeneous chart types such as dashboards, infographics, maps, bar charts, and line charts, substantially increasing both visual and linguistic diversity over earlier chart-question-answering datasets.

\textbf{SkillsBench} \cite{li2026skillsbenchbenchmarkingagentskills} evaluates whether structured packages of procedural knowledge improve the performance of language-model agents on expertise-heavy tasks. Its current inventory contains 87 tasks across eight domains, each paired with curated Skills and deterministic verifiers. The benchmark uses matched evaluations with and without Skills to separate the contribution of procedural guidance from the underlying model and agent harness. We evaluated SkillsBench on OpenClaw 2026.5.7.

\textbf{Terminal-Bench 2.1} \cite{merrill2026terminalbenchbenchmarkingagentshard,terminalbench2026v21} evaluates agents on 89 difficult, realistic tasks executed in isolated command-line environments across software engineering, machine learning, security, data science, and related workflows. Each task provides a dedicated environment, a human-written reference solution, and automated tests for outcome verification. Version 2.1 revises 28 tasks from Terminal-Bench 2.0 and introduces continuous validation to improve task correctness and benchmark reliability. We evaluated Terminal-Bench 2.1 on Terminus 2, and some results were collected from Artificial Analysis.

\textbf{SWE-Bench Pro} \cite{deng2025swebenchproaiagents} evaluates coding agents on long-horizon, enterprise-oriented software-engineering tasks that may require hours or days of professional work. It contains 1,865 human-verified problems from 41 actively maintained repositories, divided into public, held-out, and commercial sets spanning open-source and proprietary codebases. Agents must implement substantial, often multi-file changes that satisfy task-specific tests without regressing existing behavior. We evaluated SWE-Bench Pro on Mini-SWE-Agent, and we modified the official evaluation image to avoid agents getting ground truth from git logs.

\textbf{SWE-bench Multilingual} \cite{khandpur2025swebenchmultilingual,yang2025swesmith} extends SWE-bench-style issue resolution beyond Python with 300 manually curated tasks from 42 repositories and nine programming languages: C, C++, Go, Java, JavaScript, TypeScript, PHP, Ruby, and Rust. Each task provides a real GitHub issue and a pre-solution repository snapshot, requiring an agent to generate a patch that passes both fail-to-pass tests for the requested fix and pass-to-pass tests for existing functionality. It remains compatible with the standard SWE-bench evaluation infrastructure while exposing language-specific differences in software-engineering performance. We evaluated SWE-Bench Multilingual on Mini-SWE-Agent.

\textbf{WildClawBench}~\cite{ding2026wildclawbench} is a native-runtime benchmark designed to evaluate autonomous AI agents on complex, long-horizon tasks within real CLI harness environments. It integrates human-authored bilingual and multimodal workflows with real tools and environments to test agents' end-to-end tool orchestration and execution capabilities.

\begin{table}[t]
\centering
\caption{Comprehensive performance comparison across scientific benchmarks. \open{Underline} means the best performance among open-sourced models, \best{bold} indicates the best performance among all models.
}
\label{tab:scientific_results}
\normalsize
\begin{adjustbox}{max totalsize={\textwidth}{\textheight}}
\begin{tabular}{
  L{3cm}
  D{5cm}
  B{2.4cm}
  C{1.9cm}
  C{1.9cm}
  C{1.9cm}
  C{1.9cm}
  C{1.9cm}
  C{1.9cm}
  C{1.9cm}
}
\multicolumn{10}{c}{}\\[-0.4em]
\SubTableTitle{Scientific Tasks}
\toprule
\textbf{Benchmark} & \textbf{Description} &
\modelhead{Intern-S2-}{Preview-397B}{} &
\modelhead{Qwen3.5-}{397B-A17B}{} &
\modelhead{DeepSeek-}{V4-pro}{} &
\modelhead{Kimi-}{K2.7-Code}{} &
\modelhead{GLM-}{5.2}{} &
\modelhead{GPT-}{5.5}{} &
\modelhead{Gemini-}{3.1-Pro}{} &
\modelhead{Claude-}{Opus-4.8}{} \\
\midrule

Biology-Instructions & Multi-omics Sequence Analysis & \best{\open{56.92}} & 4.49 & 9.14 & 7.68 & 6.34 & 10.52 & 13.87 & 6.78 \\
Mol-Instructions & Bio-molecular Instruction & \best{\open{52.37}} & 11.65 & 12.06 & 24.56 & 19.58 & 40.49 & 38.84 & 38.35 \\
MolecularIQ & Molecular Structure Reasoning & \open{61.49} & 41.48 & 44.43 & 52.81 & 60.91 & \best{76.41} & 38.94 & 66.78 \\
SciReasoner & Scientific Reasoning & \best{\open{63.97}} & 45.02 & 51.11 & 51.69 & 51.45 & 61.15 & 60.35 & 58.00 \\
TOMG-Bench & Molecule Generation & \open{65.66} & 54.06 & 57.63 & 58.28 & 57.89 & \best{69.89} & 62.67 & 61.38 \\
MP20 & Material Structure Generation & \best{\open{67.88}} & 6.15 & 6.75 & 8.40 & 1.50 & 16.12 & 16.75 & 15.60 \\
ProteinBinder-9 & Biomolecular Interaction Design & \best{\open{4.36}} & 1.64 & 1.88 & 1.92 & 2.01 & 2.13 & 2.21 & 2.40 \\
\midrule
\multicolumn{10}{l}{\cellcolor{subgray}\hspace{-2pt}\textbf{MultiModal Tasks}}\\
XLRS-Bench & Remote Sensing & \open{51.97} & 50.11 & -- & 49.90 & -- & 50.96 & \best{54.27} & 51.84 \\
MicroVQA & Biological Microscopy VQA & \open{68.81} & 68.71 & -- & 61.04 & -- & 63.63 & \best{71.02} & 61.80 \\
SFE & Scientific Multimodal Tasks & 61.67 & \best{\open{62.97}} & -- & 50.76 & -- & 52.09 & 59.57 & 59.08 \\
ObsCrisis-Bench & Extreme Weather Analysis & 26.07 & 19.22 & -- & \best{\open{32.63}} & -- & 28.33 & 25.71 & 24.24 \\
\midrule
\multicolumn{10}{l}{\cellcolor{subgray}\hspace{-2pt}\textbf{Agentic Tasks}}\\
SciCode & Agentic Scientific Coding & 49.11 & 46.35 & 47.53 & 43.49 & \open{51.97} & 55.92 & 54.44 & \best{56.21} \\
SGI-Bench & Scientific Agent Interaction & 49.37 & 44.44 & 45.70 & 50.63 & \best{\open{52.41}} & 42.77 & 45.28 & 49.06 \\
ResearchClawBench & Automated Research & 18.44 & 15.86 & 13.69 & 15.40 & \best{\open{23.35}} & 17.00 & 14.54 & 21.74 \\

\bottomrule
\end{tabular}
\end{adjustbox}
\end{table}

\begin{table}[t]
\centering
\caption{Comprehensive performance comparison across general benchmarks. \open{Underline} means the best performance among open-sourced models, \best{bold} indicates the best performance among all models.}
\label{tab:general_results}
\normalsize
\begin{adjustbox}{max totalsize={\textwidth}{\textheight}}
\begin{tabular}{
  L{3cm}
  D{5cm}
  B{2.4cm}
  C{1.9cm}
  C{1.9cm}
  C{1.9cm}
  C{1.9cm}
  C{1.9cm}
  C{1.9cm}
  C{1.9cm}
}
\multicolumn{10}{c}{}\\[-0.4em]
\SubTableTitle{General Tasks}
\toprule
\textbf{Benchmark} & \textbf{Description} &
\modelhead{Intern-S2-}{Preview-397B}{} &
\modelhead{Qwen3.5-}{397B-A17B}{} &
\modelhead{DeepSeek-}{V4-pro}{} &
\modelhead{Kimi-}{K2.7-Code}{} &
\modelhead{GLM-}{5.2}{} &
\modelhead{GPT-}{5.5}{} &
\modelhead{Gemini-}{3.1-Pro}{} &
\modelhead{Claude-}{Opus-4.8}{} \\
\midrule

MMLU Pro & Knowledge \& Reasoning & \open{89.75} & 87.80 & 86.86 & 87.10 & 87.22 & 88.20 & \best{91.00} & 90.12 \\
SimpleQA-Verified & Factual Question Answering & \open{69.90} & 54.80 & 46.60 & 38.60 & 37.90 & 64.30 & \best{75.60} & 43.30 \\
AdvancedIF & Instruction Following & 74.44 & 75.49 & 73.83 & \open{76.17} & 75.76 & 76.20 & \best{79.78} & 72.88 \\
HMMT-2026 & High School Mathematics Competition & 91.57 & 87.88 & 91.76 & 90.34 & \open{92.50} & \best{97.06} & 94.70 & 95.36 \\
\midrule
\multicolumn{10}{l}{\cellcolor{subgray}\hspace{-2pt}\textbf{MultiModal Tasks}}\\
MMMU Pro & Knowledge \& Reasoning & \open{80.46} & 80.29 & -- & 77.92 & -- & 81.68 & \best{83.99} & 76.88 \\
ChartQAPro & Chart Question Answering & \open{69.65} & 68.61 & -- & 54.86 & -- & 69.23 & \best{71.18} & 58.65 \\
\midrule
\multicolumn{10}{l}{\cellcolor{subgray}\hspace{-2pt}\textbf{Agentic Tasks}}\\
SkillsBench & Skill usage in Harness & 50.03 & 35.58 & 49.53 & \best{\open{55.63}} & 53.19 & 49.59 & 37.20 & 54.40 \\
TerminalBench 2.1 & Terminal Mastery & 67.42 & 51.30 & 64.00 & 66.29 & \open{77.90} & 79.40 & 73.80 & \best{84.60} \\
SWE-Bench-Pro & Software Engineering & 61.56 & 43.55 & 55.40 & 57.59 & \open{62.10} & 58.60 & 54.20 & \best{69.20} \\
SWE-Bench-Multilingual & Software Engineering & 81.67 & 65.00 & 72.44 & 78.56 & \best{\open{82.00}} & 73.33 & 44.00 & 77.00 \\
WildClawBench & Real-World Agent Tasks & 44.68 & 34.50 & 43.70 & 46.89 & \open{54.20} & 58.20 & 40.80 & \best{64.72} \\
\bottomrule
\end{tabular}
\end{adjustbox}

\end{table}

\subsection{Main Results} 

As shown in Table~\ref{tab:scientific_results} and Table~\ref{tab:general_results}, Intern-S2-Preview-397B demonstrates leading performance across a broad range of scientific benchmarks. It outperforms strong open- and closed-source models on Biology-Instructions (56.92), Mol-Instructions (52.37), and SciReasoner (63.97). The model also achieves state-of-the-art (SOTA) results on our internal MP20 and ProteinBinder-9 evaluation sets. Furthermore, it delivers the best performance among open-source models on MolecularIQ (61.49), TOMG-Bench (65.66), XLRS-Bench (51.97), and MicroVQA (68.81).

On science-oriented agentic tasks, Intern-S2-Preview-397B generally surpasses DeepSeek-V4-Pro and Qwen3.5-397B, ranking second only to GLM-5.2. The model also performs strongly on general-purpose benchmarks, achieving the best results among open-source models on MMLU-Pro (89.75), SimpleQA-Verified (69.90), MMMU-Pro (80.46), and ChartQAPro (69.65). On general-purpose agentic tasks, it consistently outperforms Qwen3.5-397B and demonstrates performance comparable to that of Kimi-K2.7-Code.

\subsection{Study of Architectures}
\label{sec:study-of-architectures}

\begin{figure}[t]
\centering

\begin{minipage}[t][0.515\textheight][s]{0.54\linewidth}
\vspace{0pt}
\centering
{%
\footnotesize
\setlength{\tabcolsep}{2.5pt}
\renewcommand{\arraystretch}{1.28}
\begin{tabularx}{\linewidth}{
  @{}
  >{\raggedright\arraybackslash}X
  >{\centering\arraybackslash}p{0.22\linewidth}
  >{\centering\arraybackslash}p{0.27\linewidth}
  @{}
}
\toprule
& \multicolumn{2}{c@{}}{\textbf{Intern-S2-Preview-397B}} \\
\cmidrule(l){2-3}
\textbf{BioIns Task}
& \textbf{w/o MemDec}
& \textbf{w/ MemDec-4B} \\
\midrule
DNA-cpd                                      & 63.11 & 72.57 \\
DNA-emp                                      & 19.95 & 27.25 \\
DNA-enhancer activity                        & 53.68 & 60.71 \\
DNA-pd                                       & 84.40 & 89.12 \\
DNA-tf-h                                     & 56.57 & 55.99 \\
DNA-tf-m                                     & 56.96 & 67.09 \\
Multi-sequence antibody-antigen              & 40.24 & 36.44 \\
Multi-sequence promoter-enhancer interaction & 22.46 & 38.47 \\
Multi-sequence RNA-protein interaction       & 84.74 & 87.34 \\
Multi-sequence siRNA efficiency              & 63.05 & 60.63 \\
Protein-Fluorescence                         & 70.48 & 72.23 \\
Protein-FunctionEC                           & 61.88 & 60.10 \\
Protein-Solubility                           & 68.60 & 68.00 \\
Protein-Stability                            & 69.67 & 67.80 \\
Protein-Thermostability                      & 58.44 & 53.97 \\
RNA-CRISPROnTarget                           &  6.61 & 17.18 \\
RNA-Isoform                                  & 82.65 & 84.81 \\
RNA-MeanRibosomeLoading                      & 56.20 & 59.71 \\
RNA-Modification                             & 59.64 & 60.48 \\
RNA-NoncodingRNAFamily                       & 78.80 & 85.70 \\
RNA-ProgrammableRNA Switches                 & 37.13 & 41.23 \\
\midrule
\textbf{AVG score} & \textbf{56.92} & \textbf{60.32} \\
\bottomrule
\end{tabularx}
}
\vfill
{\footnotesize\textbf{(a)} Biology-Instructions task-level results}
\end{minipage}
\hfill
\begin{minipage}[t][0.515\textheight][s]{0.43\linewidth}
\vspace{0pt}
\centering
\includegraphics[width=0.96\linewidth]
  {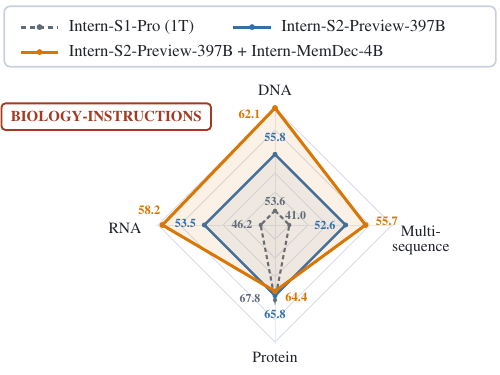}
\par\vspace{2pt}
{\footnotesize\textbf{(b)} Biology-Instructions category radar plot}

\vfill

\includegraphics[width=0.98\linewidth]
  {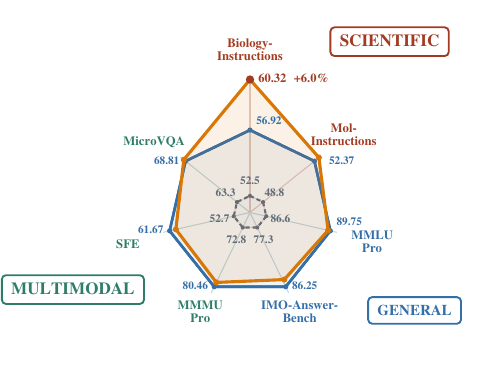}
\par\vspace{2pt}
{\footnotesize\textbf{(c)} Cross-domain capability profile}
\end{minipage}

\caption{
Evaluation of the separate Memory Decoder extension.
(a) Results on all 21 Biology-Instructions
tasks~\cite{he2025biology}.
(b) Radar plot of category averages on task-specific normalized
axes, with Intern-S2-Preview-397B as the reference square.
(c) Radar plot comparing Intern-S1-Pro (1T),
Intern-S2-Preview-397B, and Intern-S2-Preview-397B with Intern-MemDec-4B across seven
benchmarks. Colors and boxed labels denote task families.
}
\label{fig:memory-decoder-evaluation}
\end{figure}

\paragraph{Memory Decoder.}\label{sec:memory-decoder-results} To evaluate Memory Decoder as a separate extension of Intern-S2-Preview-397B, we select biology as a representative scientific domain and instantiate Intern-MemDec-4B. We evaluate the resulting biology memory on all 21 Biology-Instructions tasks~\cite{he2025biology} and examine cross-domain behavior on MMLU Pro, Mol-Instructions, MMMU Pro, MicroVQA, IMO-Answer-Bench, and SFE, with Intern-S1-Pro (1T) included as an additional reference model.
As shown in Figure~\ref{fig:memory-decoder-evaluation}, the memory-augmented variant improves the Biology-Instructions average score from 56.92 to 60.32 relative to the frozen Intern-S2-Preview-397B backbone. The cross-domain profile is used to check whether the biology memory changes behavior outside the target domain. In these comparisons, Intern-MemDec-4B remains close to the frozen backbone on general knowledge, reasoning, scientific, and multimodal benchmarks while improving the target biology benchmark, indicating that Memory Decoder can provide targeted scientific specialization as an optional extension of the 397B backbone.

\begin{table}[t]
\centering
\caption{Results of time series understanding on SciTS benchmark. F1 scores are reported. Higher F1 indicates better performance.}
\label{tab: res}
\setlength{\tabcolsep}{3pt}
\resizebox{\linewidth}{!}{
\begin{tabular}{@{}llccccccccccc@{}}
\toprule
 & \textbf{SciTS Task ID} &  \textbf{ASU01} & \textbf{ASU03} & \textbf{BIU01} & \textbf{BIU03} & \textbf{EAU01} & \textbf{MEU01} & \textbf{NEU06} & \textbf{PHU01} & \textbf{PHU04} & \textbf{RAU01} & \textbf{RAU02}\\
\midrule
\multirow{3}{*}{Text LLM} & GPT-4.1-mini        & 67.2 & 15.6 & 0.2  & 12.7 & 67.0 & 44.0 & 16.1 & 24.0 & 52.7 & 24.6 & 10.6\\
                           & Gemini2.5-Flash     & 64.1 & 16.3 & 1.5  & 12.4 & 67.6 & 60.9 & 5.8  & 20.7 & 64.8 & 20.9 & \underline{13.5}\\
                           & DeepSeek-V3         & 1.1  & 12.3 & 0.0  & 5.8  & 40.2 & 59.3 & 13.6 & 28.9 & 50.7 & 19.4& 4.2\\
\midrule
\multirow{2}{*}{VL LLM}   & GPT-5-mini          & 65.7 & 18.9 & 0.8  & 17.9 & 67.6 & 30.4 & 13.3 & 21.4 & 47.8 & 24.3 & 9.1 \\
                           & Gemini2.5-Flash     & 61.6 & 15.2 & 0.9  & 8.3  & 72.5 & 64.1 & 11.6 & 22.7 & 59.0 & \underline{31.6} & 11.3 \\
                           \midrule
                           & Intern-S1-Pro       & \textbf{98.0} & \underline{75.9} & \underline{20.8} & \underline{88.3} & \underline{99.5} & \underline{65.6} & \textbf{71.3} & \underline{36.8} & \underline{93.2} & - & - \\
                           \rowcolor{internblue} & Intern-S2-Preview-397B   & \underline{97.1} & \textbf{91.0} & \textbf{36.5} & \textbf{98.3} & \textbf{100.0} & \textbf{81.8} & \underline{70.2} & \textbf{66.9} & \textbf{99.9} & \textbf{88.4} & \textbf{60.2} \\
\bottomrule
\end{tabular}
}
\end{table}

\paragraph{Time Series Understanding.} Table~\ref{tab: res} reports the performance of Intern-S2-Preview-397B on the time series understanding tasks of the SciTS benchmark~\citep{wu2025scits}. Intern-S2-Preview consistently outperforms general-purpose Text LLMs and Vision-Language LLMs, highlighting the importance of directly modelling the underlying time series rather than relying solely on textual descriptions or visualized signals. More importantly, Intern-S2-Preview-397B achieves comparable or even better performance with less than half the number of parameters of the trillion-parameter-scale Intern-S1-Pro, surpassing it on seven of the nine tasks supported by both models. The improvements are particularly pronounced on ASU03, BIU01, BIU03, MEU01, and PHU01, with the F1 score on PHU01 increasing from 36.8 to 66.9.
The upgraded time series module further extends the model to radar coding-scheme classification and mode-and-modulation classification, which was not supported by Intern-S1-Pro, and achieves substantially better performance than the baseline models on both tasks.

\begin{table*}[t]
\centering
\caption{Results of time series forecasting on the SciTS benchmark, reported in the format MAPE (success rate \%).
Lower MAPE indicates better performance, while higher success rate is better.}
\label{tab:forecasting_results}

\begingroup

\newcommand{\mapsr}[2]{#1\,{\footnotesize(#2)}}

\resizebox{\linewidth}{!}{
\begin{tabular}{llccccccc}
\toprule
& \textbf{SciTS Task ID}
& \textbf{ENG02}
& \textbf{ENG03}
& \textbf{MEG03}
& \textbf{NEG03}
& \textbf{PHG02}
& \textbf{URG01}
& \textbf{URG05} \\
\midrule

\multirow{3}{*}{\makecell{Text\\LLM}}
& GPT-4.1-mini
& \mapsr{125.0}{1.4}
& \mapsr{8.3}{96.0}
& \mapsr{42.1}{49.6}
& \mapsr{95.2}{96.4}
& \mapsr{1.1e3}{94.2}
& \mapsr{320.6}{18.6}
& \mapsr{126.6}{100} \\

& Gemini2.5-Flash
& \mapsr{72.5}{5.9}
& \mapsr{9.6}{99.0}
& \mapsr{62.2}{57.9}
& \mapsr{63.5}{99.2}
& \mapsr{110.8}{99.0}
& \mapsr{246.0}{23.3}
& \mapsr{98.6}{100} \\

& DeepSeek-V3
& \mapsr{117.2}{46.1}
& \mapsr{7.7}{98.0}
& \mapsr{46.4}{30.9}
& \mapsr{4.3}{3.1}
& \mapsr{200.1}{92.2}
& \mapsr{350.0}{18.6}
& \mapsr{296.7}{93.0} \\

\midrule

\multirow{2}{*}{\makecell{VL\\LLM}}
& GPT-5-mini
& \mapsr{56.1}{4.5}
& \mapsr{11.2}{76.0}
& \mapsr{37.6}{51.8}
& \mapsr{74.3}{97.2}
& \mapsr{155.3}{97.4}
& \mapsr{182.1}{58.1}
& \mapsr{71.1}{72.9} \\

& Gemini2.5-Flash
& \mapsr{103.9}{7.4}
& \mapsr{15.6}{53.0}
& \mapsr{53.1}{37.2}
& --
& \mapsr{185.2}{36.9}
& \mapsr{351.9}{16.3}
& \mapsr{114.6}{91.2} \\

\midrule

\multirow{5}{*}{\makecell{Time\\Series\\Models}}
& Moirai-Large~\cite{woo2024moirai}
& \mapsr{121.2}{100}
& \mapsr{12.8}{100}
& \mapsr{51.7}{100}
& \mapsr{59.1}{100}
& \mapsr{116.9}{100}
& \mapsr{294.7}{100}
& \mapsr{74.6}{100} \\

& TimeMoE-Large~\cite{shi2025timemoe}
& \mapsr{70.4}{100}
& \mapsr{11.6}{100}
& \mapsr{39.0}{100}
& \mapsr{70.1}{100}
& \mapsr{80.2}{100}
& \mapsr{218.4}{100}
& \mapsr{84.4}{100} \\

& Chronos-bolt-Base~\cite{ansari2024chronos}
& \mapsr{73.7}{100}
& \mapsr{12.0}{100}
& \mapsr{41.5}{100}
& \mapsr{78.5}{100}
& \mapsr{109.3}{100}
& \mapsr{139.3}{100}
& \mapsr{70.6}{100} \\

& UniTS~\cite{gao2024units}
& \mapsr{70.1}{100}
& \mapsr{12.8}{100}
& \mapsr{42.0}{100}
& \mapsr{95.2}{46.4}
& \mapsr{135.9}{44.1}
& \mapsr{389.7}{100}
& -- \\

& TimeOmni~\cite{wu2025scits}
& \mapsr{68.6}{100}
& \mapsr{7.4}{100}
& \mapsr{37.5}{100}
& \mapsr{78.7}{100}
& \mapsr{163.0}{100}
& \mapsr{247.0}{100}
& \mapsr{174.0}{100} \\

\midrule

\rowcolor{internblue} & Intern-S2-Preview-397B
& \mapsr{\textbf{60.2}}{100}
& \mapsr{\textbf{7.1}}{100}
& \mapsr{\textbf{32.8}}{100}
& \mapsr{\textbf{59.2}}{100}
& \mapsr{\textbf{72.2}}{100}
& \mapsr{\textbf{138.9}}{100}
& \mapsr{\textbf{60.6}}{100} \\

\bottomrule
\end{tabular}
}
\endgroup
\end{table*}

\paragraph{Time Series Generation.} Table~\ref{tab:forecasting_results} benchmarks Intern-S2-Preview against general-purpose Text LLMs, Vision-Language LLMs, and specialised time series forecasting models on the forecasting tasks of SciTS~\citep{wu2025scits}. Text and Vision-Language LLMs often exhibit low success rates because long prediction horizons can exceed their output capacity, while strict sequence-length and formatting requirements frequently lead to instruction-following failures. Moreover, generating forecasts through discrete text tokens can compromise numerical precision, limiting their accuracy on fine-grained scientific signals. By employing a dedicated numerical forecasting branch, Intern-S2-Preview-397B preserves numerical fidelity while enabling reliable forecasting across heterogeneous domains and prediction horizons, outperforming the specialised time series baselines with particularly clear gains on ENG02, ENG03, MEG03, PHG02, and URG05. The horizon predictor achieves an accuracy of 99\%, demonstrating its ability to reliably infer the required prediction length from forecasting instructions. Beyond the scientific time series tasks in SciTS, we further evaluate Intern-S2-Preview-397B on GIFT-Eval~\citep{aksu2024gifteval}, a benchmark for general time series forecasting, where it achieves a competitive zero-shot MASE of 0.785.

%% file: sections/6.conclusion.tex
\section{Conclusion}

We presented Intern-S2-Preview-397B as a scientific agentic foundation model for scientific research that requires multimodal understanding, domain-specific reasoning, scientific generation, tool interaction, and iterative execution. Across scientific, multimodal, agentic, general-purpose, and time-series evaluations, the model demonstrates broad capability coverage and supports the main design choices in architecture, pre-training, and post-training. The agentic evaluations further indicate that scientific capability should be assessed not only by isolated benchmark-answer accuracy, but also by whether a model can connect reasoning to executable, verifiable, and iterative workflows. Separately, the Memory Decoder study shows that targeted scientific specialization can be added to the frozen backbone while preserving the role of Intern-S2-Preview-397B as the general model. Intern-S2-Preview remains a preview system; future work should improve reliability over longer scientific workflows, expand domain-specific memories and task environments, strengthen verifiers, and deepen integration with specialized scientific tools.

\subsubsection*{Author Contributions}
The authors are listed in alphabetical order by their last names.

Lei Bai, Jiaqi Cao, Chiyu Chen, Guanzhou Chen, Kai Chen, Guangran Cheng, Erfei Cui, Xuanlang Dai, Shengyuan Ding, Shangheng Du, Yanhui Duan, Yue Fan, Youqing Fang, Quan Gan, Yuanyuan Gao, Jiaye Ge, Lixin Gu, Yuzhe Gu, Qipeng Guo, Junjun He, Xin Hong, Ming Hu, Zhouqi Hua, Haian Huang, Junhao Huang, Zixian Huang, Minxi Jin, Lingkai Kong, Alexander Lam, Zehao Li, Zonglin Li, Tianhao Liang, Dahua Lin, Junyao Lin, Tianyang Lin, Zhouhan Lin, Jiangning Liu, Jin Liu, Kuikun Liu, Wenran Liu, Yifei Liu, Yuhong Liu, Yuhong Liu, Zhoumianze Liu, Ziyan Liu, Ziyu Liu, Haijun Lv, Han Lv, Chengqi Lyu, Le Ma, Ningsheng Ma, Zerun Ma, Haoyang Peng, Runyu Peng, Jifei Shan, Zixin Shang, Kou Shi, Xiang Shi, Qisheng Su, Xuerui Su, Hao Sun, Xiao Sun, Yanan Sun, Yu Sun, Huanze Tang, Yinghao Tang, Wenhui Tian, Zhongbo Tian, Bingli Wang, Haomin Wang, Jiarui Wang, Jingzhi Wang, Rui Wang, Xiquan Wang, Yi Wang, Zhecan Wang, Ziyi Wang, Zun Wang, Rubin Wei, Lianyi Wu, Wen Wu, Yue Wu, Yuhan Wu, Zhenyu Wu, Zijian Wu, Shuhao Xing, Jun Xu, Xingle Xu, Xuenan Xu, Xiangchao Yan, Ziang Yan, Bowen Yang, Danni Yang, Lin Yang, Zhiqi Yang, Qian Yao, Haochen Ye, Peng Ye, Jinhui Yin, Jiashuo Yu, Dingbo Yuan, Fei Yuan, Yuhang Zang, Bo Zhang, Chao Zhang, Chen Zhang, Hongjie Zhang, Junming Zhang, Wenlong Zhang, Wenwei Zhang, Yiming Zhang, Zhuo Zhang, Ziyang Zhang, Haiteng Zhao, Penghao Zhao, Yibo Zhao, Zhonghan Zhao, Zhihang Zhong, Bowen Zhou, Peiheng Zhou, Xin Zhou, Xinyu Zhou, Yunhua Zhou, Dongsheng Zhu, Yicheng Zou